\documentclass{article} % For LaTeX2e
\usepackage{iclr2027_conference,times}
\usepackage{xcolor}
\usepackage{amsmath,amsfonts,bm}

\def\eqref#1{equation~\ref{#1}}
\def\1{\bm{1}}

\DeclareMathAlphabet{\mathsfit}{\encodingdefault}{\sfdefault}{m}{sl}
\SetMathAlphabet{\mathsfit}{bold}{\encodingdefault}{\sfdefault}{bx}{n}

\DeclareMathOperator*{\argmax}{arg\,max}

\usepackage{hyperref}
\usepackage{url}

\usepackage{graphicx}
\graphicspath{{figures/}{./}}
\usepackage{wrapfig}
\usepackage{amsmath,amssymb,amsthm}
\usepackage[capitalise]{cleveref}
\usepackage{booktabs}

\theoremstyle{plain}
\newtheorem{lemma}{Lemma}
\newtheorem{theorem}[lemma]{Theorem}
\newtheorem{corollary}[lemma]{Corollary}
\newtheorem{proposition}[lemma]{Proposition}

\title{A Boundary-Layer Mechanism for One-Third Scaling in Online Softmax Classification}

\author{%
Marcel K\"uhn$^{1,2,}$\thanks{Corresponding author: \texttt{mkuehn@itp.uni-leipzig.de}} \quad Yoon Thelge$^1$ \quad Bernd Rosenow$^{1,2}$ \\
$^1$Institute for Theoretical Physics, Leipzig University\\
$^2$ScaDS.AI Dresden/Leipzig\\
}
\iclrfinalcopy % Uncomment for camera-ready version, but NOT for submission.
\begin{document}

%%%%%%%%%%%%%%%%%%%%%%%%%%%%%%%%%%%%%%%%%
%%%%%%%%%%%%%%%%%%%%%%%%%%%%%%%%%%%%%%%%%
%%%%%%%%%%%%%%%%%%%%%%%%%%%%%%%%%%%%%%%%%

\maketitle

\begin{abstract}

{Hard-label classification is usually trained with smooth surrogate losses such as softmax cross-entropy. We isolate a mechanism by which this mismatch between smooth surrogate and discrete labels produces power-law learning curves for a one-layer softmax student trained by online SGD on Gaussian inputs with argmax teacher labels. In the thermodynamic limit, the dynamics reduce to a growing student-teacher alignment and a residual student variance sustained by SGD noise. At late times, only shrinking layers of examples around teacher decision boundaries remain active. This yields an $\alpha^{-1/3}$ law in training time $\alpha$ not only for the test loss but also for the generalization error, one minus test accuracy. The error law is an online-noise effect. The loss law persists without gradient noise. Learning-rate annealing separates the two exponents, improving the error exponent from 1/3 toward 1/2 while slowing the loss. Simulations support the predictions. Correlated Gaussian inputs and whitened pretrained features change transients, not the late-time decay. Experiments with deep neural networks can show compatible error scaling. The mechanism is asymptotic and complementary to spectral explanations and data-statistics explanations of neural scaling laws.}
\end{abstract}

\section{Introduction}
Scaling laws make learning dynamics predictable: they summarize how error decreases with data, compute, model size, or optimization time. Empirical work has made such laws central to modern machine learning \citep{hestness2017deep,kaplan2020scaling,rae2021scaling,hoffmann2022chinchilla}.
A major theoretical route explains these laws from structure in the data distribution, target function, or feature map. In kernel, random-feature, and linear models, power-law spectra of feature or data covariance matrices can be converted into power-law learning curves \citep{bordelon2020spectrum,canatar2021spectral,maloney2022solvable,bahri2024explaining,lin2024scalinglinear}; 
related geometric considerations connect exponents to the effective dimension of the data manifold \citep{nakada2020adaptive,sharma2022scaling}. Feature-learning analyses show how these spectral mechanisms can change once the representation evolves during training \citep{bordelon2024dynamical,bordelon2025featurelearning,worschech2025analyzing}.

Power laws can also arise from mechanisms that are not primarily spectral. Simplified sequence-modeling models show how scaling behavior can emerge without explicit power-law correlations in the data \citep{barkeshli2026origin}, while recent work shows that softmax and cross-entropy can intrinsically produce one-third time scaling when learning peaked probability distributions \citep{liu2026universal}. High-dimensional teacher--student theories provide another complementary viewpoint, with precise learning curves for multiclass classification under Bayes-optimal and empirical-risk minimization procedures, including cross-entropy loss \citep{cornacchia2023learning}. Here we isolate a different late-time mechanism: a leading-order boundary-layer realization of the softmax/cross-entropy bottleneck in online multiclass classification, related to earlier observations that late-time updates concentrate near decision boundaries \citep{paccolat2021geometric,sclocchi2024different}.

This motivates a focused question: what controls the late-time learning dynamics of online multiclass classification when the labels are discrete but the student is trained through differentiable probabilities? We study this question in a one-layer $K$-class teacher-student model with Gaussian inputs and online stochastic gradient descent (SGD), using the order-parameter methodology of the statistical physics of learning \citep{opper1991calculation,biehl1994online,saad1995exact}. The model is simple enough to solve in the thermodynamic limit, but rich enough to expose an asymptotic mechanism that is easy to miss in the raw order parameters.

%%%%%%%%%%%%%%%%%%%%%%%%%%%%%%%%%%%%%%%%

\paragraph{Our contributions.}

\begin{figure}[t]
  \centering
  \includegraphics{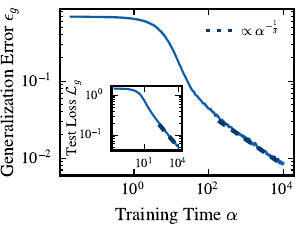}
  \hfill
  \includegraphics{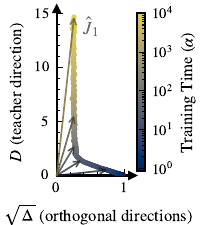}
  \hfill
  \includegraphics{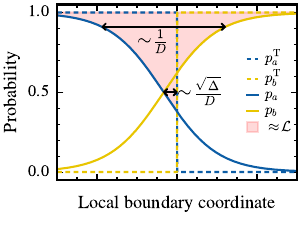}
   \caption{Left: The $1/3$ law appears not only in the test loss but also in the generalization error $\epsilon_g$, i.e., one minus test accuracy. Middle: The model captures both the growth of the centered student-teacher alignment $D$ and the rotational alignment to the teacher. Right: Near a teacher decision boundary, the late-time loss is controlled by the student boundary layer of width $O(D^{-1})$. The generalization error is controlled by residual boundary fluctuations only of order $\sqrt{\Delta}/D$.}
  \label{fig:concept}
\end{figure}

We derive an exact centered macroscopic closure for symmetric online $K$-class softmax learning. We then give a boundary-layer derivation of the softmax/cross-entropy bottleneck in hard-label classification, compute the leading classification-error asymptotics, and show how a learning-rate schedule can improve them. 
The results are stated formally under explicit assumptions (A1)--(A7), as \cref{lem:closure,lem:entry}, Theorem \ref{thm:blexp}, and Corollaries \ref{cor:fixed} and \ref{cor:sched}.
See \cref{fig:concept} for an overview.
{We test the mechanism with finite-dimensional simulations and controlled departures from isotropic Gaussian inputs, analyze exponent masking and recovery in a Zipf model, and examine Pythia token top-1 error,  finding a fitted exponent compatible with $1/3$ for the $12$B model.}

The late-time dynamics becomes transparent only in centered coordinates, which remove the softmax redundancy under a common shift of all logits (\cref{sec:centered-dynamics}): the centered student-teacher overlap $D$, the centered student squared norm $Q_{\mathrm{eff}}$, and the residual student variance
$\Delta = Q_{\mathrm{eff}} - D^2$,
which comes from fluctuations of the student orthogonal to the teacher. In these variables, perfect learning is not convergence to a finite-weight fixed point. Instead, the centered overlap $D$ grows without bound, while at fixed learning rate the residual variance $\Delta$ approaches a finite noise floor. The only examples that remain active lie within $O(D^{-1})$ of pairwise teacher decision boundaries. The shrinking measure of these active boundary layers gives $\dot D\propto D^{-2}$, and therefore $D\propto \alpha^{1/3}$. The classification error is controlled by the angle between student and teacher, $\epsilon_g\propto \sqrt{\Delta}/D$, giving
\begin{equation}
  \epsilon_g(\alpha)\propto \alpha^{-1/3} \ .
\end{equation}
Annealing the learning rate changes the residual noise floor. For a slowly decaying learning rate $\eta(\alpha)\propto \alpha^{-\gamma}$ with $0\leq \gamma<1$, this gives $\epsilon_g(\alpha)\propto \alpha^{-(2+\gamma)/6}$, improving toward $\epsilon_g \propto \alpha^{-1/2}$ as $\gamma \to 1$.
This rate remains slower than Bayes-optimal rates in theories, but mirrors reports for cross-entropy minimization at fixed regularization on static datasets \citep{cornacchia2023learning}. 

These results should be read in the context of \citet{liu2026universal}, {who show that softmax and cross-entropy can generate one-third scaling of the test loss when learning peaked distributions.
Our solvable model of online hard-label classification yields classification-error asymptotics,
identifying the roles of decision-boundary geometry and the residual variance maintained by online gradient noise.}

Throughout, $x\cdot y$ is the Euclidean inner product; $f\sim g$ means $f/g\to1$ and $f\propto g$ means $c\,g\le f\le C\,g$ for constants $c,C>0$, as $\alpha\to\infty$.

\section{Related work}
\label{sec:related}

Empirical scaling laws have motivated a broad set of theoretical explanations, including spectral, geometric, and feature-learning mechanisms
\citep{bahri2024explaining,bordelon2020spectrum,canatar2021spectral,maloney2022solvable,sharma2022scaling,bordelon2024dynamical,bordelon2025featurelearning,worschech2025analyzing}.
Our work is complementary: it gives a sharply scoped classification example in which decision-boundary geometry and online optimization noise produce a power law even without a structured input spectrum.

The closest recent result is the one-third softmax and cross-entropy bottleneck of \citet{liu2026universal}. They analyze peaked target distributions under gradient flow using a solvable ansatz in which the student is exactly aligned with the teacher, so that learning changes only the overall logit scale. This yields a $1/3$ loss law, supported by evidence from large-language-model training. However, mere rescaling preserves the student logit order, so the reduced theory has identical student and teacher decision boundaries and zero classification error. It therefore does not predict how classification error decreases during learning.

For classification tasks such as image recognition, generalization error measures the decision performance of interest more directly than cross-entropy, and the two metrics need not share a scaling exponent \citep{hestness2017deep}. Exponents vary, but for ImageNet \citep{deng2009imagenet} top-1 error scaling with an exponent near one-third has been reported both for training-data scaling \citep{hestness2017deep} and saturating compute-frontier scaling \citep{zhai2022scaling}. This distinction also matters for LLMs: scaling laws have been developed for average downstream top-1 error \citep{gadre2025language}, and multiple-choice accuracy depends on the correct answer's logit relative to specific competitors, not its loss alone \citep{schaeffer2025why}. Moreover, decomposing cross-entropy loss reveals that a rank-entropy component based on the distribution of the correct token's logit rank can scale more robustly than total cross-entropy \citep{yan2026scales}. Although token top-1 error requires caution because several next tokens may be plausible, it directly probes logit ordering rather than probability magnitude.

The present hard-label setting, a limiting case of peaked targets, resolves the gradient noise and parameter rotation explicitly left open by \citet{liu2026universal} and yields the classification-error predictions of \cref{sec:boundary-layers,sec:annealing}; for the loss alone, our schedule result reduces to their time reparametrization $\tau=\int\eta$. Where their analytic derivation emphasizes softmax and cross-entropy, our boundary-layer calculation makes the geometric source of the slow drift explicit. A companion calculation for a smooth perceptron classifier trained on hard labels with mean-squared-error loss (\cref{app:binary-warmup}) illustrates the same local asymptotic mechanism beyond that surrogate. \cref{tab:liu-comparison} summarizes the comparison.

Our analysis uses the classic teacher--student and thermodynamic-limit methodology of the statistical physics of learning \citep{biehl1994online,seung1992statistical,opper1991calculation,saad1995exact}, complementary to physics-informed machine-learning methods that build physical constraints into models \citep{raissi2019physics}.
In this tradition, high-dimensional stochastic learning dynamics reduce to deterministic flows for self-averaging order parameters.
One-third exponents have been observed before, as already the standard online perceptron algorithm exhibits one-third scaling of the generalization error \citep{biehl1994online}. The present study distinguishes itself by analyzing gradient descent and multiclass classification.
Power-law late-time behavior can also arise from dynamical degeneracies such as soft modes in over-realizable soft committee machines \citep{richert2022soft}. Modern work has extended these ideas to richer neural-network and high-dimensional classification settings \citep{advani2020high,goldt2019dynamics,mignacco2020dynamical}. 

High-dimensional teacher--student analyses provide reference points for Bayes-optimal classification, convex optimization, and learning curves with generic feature maps \citep{aubin2020perceptron,loureiro2021learning,cornacchia2023learning}. These works address the asymptotic performance of estimators trained on a static set of examples, whereas the present paper addresses late-time online dynamics for hard-label classification under softmax and cross-entropy updates.

{ Closely related are analyses of hinge-loss training: for networks with one hidden layer, \citet{paccolat2021geometric} showed that examples far from the class boundaries cease to contribute to the gradient as training proceeds, so that the updates become concentrated on the examples near those boundaries. 
For a binary perceptron with Gaussian inputs, \citet{sclocchi2024different} derived a test error that decreases with a one-third power of training time, 
in an online regime in which SGD noise dominates. Here, by contrast,  we consider a multiclass classifier trained online with the standard softmax cross-entropy loss. We show that the loss and the misclassification error are controlled by two distinct boundary-layer widths, and that this distinction explains how decaying the learning rate can make the error fall faster, and the loss more slowly, than they would at a fixed learning rate.}

Our results also connect to smooth surrogate losses and implicit bias. Calibration and consistency theory relates surrogate risk to classification risk \citep{zhang2004statistical,bartlett2006convexity,wang2024unified}, and recent work gives universal square-root rates for smooth surrogate-loss $H$-consistency bounds \citep{mao2024surrogate}. 
The limiting behavior of our annealing schedule, improving toward $\epsilon_g(\alpha) \propto \alpha^{-1/2}$, echoes this square-root behavior but concerns online optimization time rather than excess-risk transforms.
For separable data, gradient descent on logistic or cross-entropy losses drives weights toward max-margin directions while the norm diverges \citep{soudry2018implicit}. Related work studies fixed-learning-rate SGD and multiclass extensions of this implicit-bias picture \citep{nacson2019stochastic,ravi2024implicit}. Our population online setting differs because fresh examples continually inject boundary noise, leaving a residual uncertainty at fixed learning rate that annealing can reduce.

\section{Centered online dynamics for the \texorpdfstring{$K$}{K}-class softmax student}
\label{sec:centered-dynamics}
%%%%%%%%%%%%%%%%%%%%%%%%%%%%%%%%%%%%%%%%%%
%%%%%%%%%%%%%%%%%%%%%%%%%%%%%%%%%%%%%%%%%%
%%%%%%%%%%%%%%%%%%%%%%%%%%%%%%%%%%%%%%%%%%

\subsection{Teacher-student model assumptions}

\emph{(A1) Single-layer student.} We study a single-layer student that maps $N$-dimensional inputs to $K$ logits, one for each class. It has weights $J_1,\ldots,J_K\in\mathbb{R}^N$, with logits and softmax probabilities
\begin{equation}
    t_a=\frac{J_a\cdot \xi}{\sqrt N},
    \qquad
    p_a=\frac{e^{t_a}}{\sum_{b=1}^K e^{t_b}} \ .
    \label{eq:student-logits}
\end{equation}
\emph{(A2) Orthonormal single-layer teacher.} Let $T_1,\ldots,T_K\in\mathbb{R}^N$ be orthogonal teacher vectors, normalized as $T_a\cdot T_b/N=\delta_{ab}$, where $\delta_{ab}$ is the Kronecker delta.

\emph{(A3) Hard labels.} Labels are generated by the teacher. The teacher fields and one-hot labels are
\begin{equation}
    u_a=\frac{T_a\cdot \xi}{\sqrt N} \ ,
    \qquad
    p_a^{\rm T}=\mathbf 1\{u_a=\max_b u_b\} \ ,
    \label{eq:teacher-model}
\end{equation}
where $\mathbf 1\{\cdot\}$ is the indicator function.  Ties have probability zero under the Gaussian input distribution. Hard labels refer to the fact that the labels are deterministic and one-hot. They form the zero-temperature limit of the peaked soft targets of \citet{liu2026universal}. 

\emph{(A4) Standard Gaussian inputs.} Inputs follow $\xi\sim\mathcal N(0,I_N)$.

\emph{(A5) Online SGD on cross-entropy.} For a single example, cross-entropy loss is $\mathcal{L}=-\log p_y$, where $y=\argmax_a u_a$ is the teacher label.
Online gradient descent with learning rate $\eta$ and fixed finite batch size $b$ gives the update
\begin{equation}
    J_a^{\mu+1}
    =
    J_a^{\mu}+\frac{\eta}{b\sqrt N}\sum_{i=1}^b\left(p_{a}^{{\rm T}\!,\mu,i}-p_a^{\mu,i}\right)\xi^{\mu,i},
    \qquad
    \alpha=\frac{\mu}{N} \ .
    \label{eq:online-update}
\end{equation}

\emph{(A6) Thermodynamic limit.} We analyze the dynamics in the limit $N\to\infty$ with $b$ and $\alpha=\mu/N$ fixed, taken before a late-time limit $\alpha\to\infty$.

\emph{(A7) Permutation-symmetric ansatz.} Under permutation symmetry between classes, the standard  statistical mechanics order parameters are
\begin{equation}
    R=\frac{J_1\cdot T_1}{N} \ ,
    \qquad
    S=\frac{J_1\cdot T_2}{N} \ ,
    \qquad
    Q=\frac{J_1\cdot J_1}{N} \ ,
    \qquad
    C=\frac{J_1\cdot J_2}{N} \ .
    \label{eq:RSQC}
\end{equation}
By symmetry, the choice of indices is arbitrary, with $1$ and $2$ denoting distinct classes. Here $R$ is the overlap with the matching teacher, $S$ is the overlap with a non-matching teacher, $Q$ is the student squared norm, and $C$ is the overlap between two distinct student weight vectors.

\paragraph{Assumption scope.} (A1), (A2), (A6), and (A7) are standard in this model class \citep{biehl1994online,saad1995exact,goldt2019dynamics,mignacco2020dynamical}: (A2) is asymptotically satisfied by independent isotropic random teachers at fixed $K$, whose normalized off-diagonal overlaps are $O(N^{-1/2})$, finite-$N$ runs at $N=200$--$500$ already match the limiting theory with (A6) at prefactor level (\cref{sec:numerics}), and the i.i.d.\ student initialization in the Gaussian teacher--student simulations self-averages to the symmetric ansatz of (A7). (A5) is standard but load-bearing: it matches the single-pass regime of large-scale training and is essential for the generalization-error results though not for the loss. (A3) and (A4) are restrictive: (A3) is essential, because soft targets or label noise can mask the boundary-layer asymptotics; (A4) is relaxed empirically in \cref{sec:numerics}.

\subsection{Exact closure in centered order parameters}

The softmax probabilities are invariant under a common shift of the logits, $t_a\mapsto t_a+c$. The mean logit is therefore dynamically irrelevant. This motivates the centered order parameters
\begin{equation}
    D:=R-S \ ,
    \qquad
    Q_{\mathrm{eff}}:=Q-C \ ,
    \qquad
    \Delta:=Q_{\mathrm{eff}}-D^2 \ .
    \label{eq:centered-variables}
\end{equation}
Their geometric meaning is transparent in centered coordinates:  define the centered student $\hat{J}_a := \sqrt{\tfrac{K}{K-1}}\Big( J_a - \bar J \Big)$ and centered teacher $\hat{T}_a := \sqrt{\tfrac{K}{K-1}}\Big( T_a - \bar T \Big)$ with $\bar T := \tfrac{1}{K}\sum_aT_a$ and $\bar J := \tfrac{1}{K}\sum_aJ_a$. 
Then, for any class $a$,
\begin{equation}
    D = \frac{\hat{J}_a\cdot \hat{T}_a}{N} \ , \qquad Q_{\rm eff} = \frac{\hat{J}_a\cdot \hat{J}_a}{N} \ .
\end{equation}
Thus $D$ is the centered student-teacher overlap and $Q_{\mathrm{eff}}$ is the centered student norm, both independent of the class index $a$ by permutation symmetry, and $\Delta$ is the residual student variance due to fluctuations of the student orthogonal to the teacher.

For the macroscopic dynamics, only the centered student logits and teacher fields matter. In the thermodynamic limit, their joint Gaussian law can be represented as
\begin{equation}
    h_a:=t_a-\bar t
    =D(u_a-\bar u)+\sqrt{\Delta}\,(z_a-\bar z) \ ,
    \label{eq:centered-logit-representation}
\end{equation}
where $\bar t=K^{-1}\sum_a t_a$, $\bar u=K^{-1}\sum_a u_a$, $\bar z=K^{-1}\sum_a z_a$, and the $z_a$ are auxiliary i.i.d.~standard Gaussians independent of the teacher fields. The first term is the centered teacher-aligned signal, with scale $D$. The second term is residual centered noise, with scale $\sqrt{\Delta}$. Since the softmax probabilities depend only on the centered logits $h_a$, this representation is used throughout the paper, simplifying the analysis by considering uncorrelated Gaussian variables $u_a$ and $z_a$.

Let  \(g_a=p^{\rm T}_a-p_a\). In the thermodynamic limit, the permutation-symmetric online learning process closes exactly on the centered variables. With dots denoting derivatives with respect to $\alpha$:

\begin{lemma}[Centered closure]\label{lem:closure}
Under (A1)--(A7), the joint Gaussian law of the centered logits in the thermodynamic limit admits the representation \cref{eq:centered-logit-representation}, and for fresh minibatches of size $b\ge1$ the online dynamics closes exactly on the centered variables:
\begin{equation}
\dot D=\frac{K}{K-1}\,\eta\,\langle g_1(u_1-\bar u)\rangle\ ,
\qquad
\dot Q_{\mathrm{eff}}=\frac{K}{K-1}\Bigl(2\eta\langle g_1h_1\rangle+\frac{\eta^2}{b}\langle g_1^2\rangle\Bigr)\ ,
\qquad
\dot\Delta=\dot Q_{\mathrm{eff}}-2D\dot D\ ,
\label{eq:closure-lemma}
\end{equation}
where the averages are over the centered Gaussian representation \cref{eq:centered-logit-representation}.
\end{lemma}

No late-time $\alpha \to \infty$ approximation has been made yet. The derivation from the microscopic update is given in \cref{app:exact-K-dynamics}.

\section{Boundary layers control the late-time regime}
\label{sec:boundary-layers}

\paragraph{Almost-perfect learning.}
The late-time regime we consider is characterized by $D\to\infty$ with bounded $\Delta$. For large $\alpha$ the dynamics of \cref{lem:closure} will enter this regime:

\begin{lemma}[Global entry into the late-time regime]\label{lem:entry}
Let $0\le\eta(\alpha)\le\eta_{\max}<\infty$ with $\int_0^\alpha\eta(s)\,ds\to\infty$. Under (A1)--(A7), for every admissible ($\Delta\ge0$) finite permutation-symmetric initial state and every fixed finite $b$,
\begin{equation}
\sup_{\alpha\ge0}\Delta(\alpha)<\infty\ ,
\qquad
D(\alpha)\to\infty\ .
\label{eq:global-entry}
\end{equation}
\end{lemma}

A proof is given in \cref{app:global-entry}. In this regime, the signal term in \cref{eq:centered-logit-representation} is large except near teacher decision boundaries: away from them, teacher and student agree exponentially well and examples contribute exponentially little to the macroscopic flow. The leading contributions eventually come from pairwise boundary layers of width $O(D^{-1})$ around $u_a=u_b$. Higher-order ties are lower-dimensional sets and are subleading. While we derive the boundary layer results below for the limit $\alpha\to \infty$ and $D \to \infty$, in \cref{app:heuristic_entry_scale} we provide a heuristic entry scale for $\alpha$ after which the boundary layer description becomes reasonable.

For a fixed pair $a\neq b$, the boundary is locally relevant only when the other $K-2$ teacher fields lie below the common value.  This gives the geometric boundary density
\begin{equation}
    c_K:=\int_{-\infty}^{\infty}\varphi(s)^2\Phi(s)^{K-2}\,ds,
    \label{eq:cK-main}
\end{equation}
where $\varphi$ and $\Phi$ are the standard normal density and distribution function. All dependence on the number of classes enters the asymptotic prefactors through this boundary density and the number of pairwise boundaries.

\paragraph{Local binary reduction.}

Near one active boundary, scale the teacher gap as
$u_a-u_b=\frac{x}{D}$.
Then the corresponding centered student-logit gap has the form
\begin{equation}
    h_a-h_b=x+\delta \ ,
    \qquad
    \delta=\sqrt{2\Delta}\,z \ ,
    \qquad
    z\sim\mathcal{N}(0,1) \ .
    \label{eq:local-gap}
\end{equation}
All other classes are exponentially suppressed at leading order, so the local softmax comparison is binary.  The universal local update is
$    \Theta(x)-\sigma(x+\delta)$ with
$      \sigma(y)=1/(1+e^{-y})$.
The only non-elementary scalar function that remains is
\begin{equation}
    \mathcal{B}(\Delta)
    =
    \int Dz\left[
    2\log\left(2\cosh\left(\sqrt{\frac{\Delta}{2}}\,z\right)\right)-1
    \right] \ ,
    \qquad
    Dz=\frac{e^{-z^2/2}}{\sqrt{2\pi}}\,dz \ .
    \label{eq:B-main}
\end{equation}
For small $\Delta$, $\mathcal{B}(\Delta)=2\log 2-1+\Delta/2+O(\Delta^2)$.

\paragraph{Uniform boundary-layer expansion.}
The classification generalization error is
\begin{equation}
    \epsilon_g
    :=
    \Pr\left[\argmax_a h_a\neq \argmax_a u_a\right].
    \label{eq:epsg-def-main}
\end{equation}
It is governed by the same boundary layers: locally, a disagreement occurs when the signs of $x$ and $x+\delta$ differ. The local reduction then gives the leading late-time behavior of all four macroscopic observables:

\begin{theorem}[Boundary-layer expansion]\label[theorem]{thm:blexp}
Let $c_K$ and $\mathcal B$ be as in \cref{eq:cK-main,eq:B-main}. Under (A1)--(A7), as $D\to\infty$, uniformly for $\Delta\in[0,M]$ with fixed $M<\infty$,
\begin{align}
    \dot D
    &=
    \frac{Kc_K\,\eta}{2D^2}
    \left(\frac{\pi^2}{6}+\Delta\right)+\eta\,o({D^{-2}}) \ ,
    \label{eq:Ddot-asymptotic-main}
    \\
    \dot\Delta
    &=
    \frac{Kc_K}{D}\left[
    \frac{\eta^2}{b}\,\mathcal{B}(\Delta)-2\eta\Delta
    \right]+\eta(\Delta+\eta/b)\,o({D^{-1}}) \ ,
    \label{eq:Deltadot-asymptotic-main}
    \\
    \epsilon_g
    &=
    \Gamma_K\frac{\sqrt{\Delta}}{D}+O\!\left(\frac{\Delta}{D^2}\right) \ ,
    \qquad
    \Gamma_K=\frac{K(K-1)c_K}{\sqrt{\pi}} \ ,
    \label{eq:epsg-boundary-main}
\end{align}
and $\mathcal L_g := \langle -\log p_y \rangle= \frac{K(K-1)c_K}{2D}\left(\frac{\pi^2}{6}+\Delta\right)+o(D^{-1})$.
\end{theorem}
The derivation is given in \cref{app:boundary-layer-derivation} with explicit remainders. The loss and the error live on the same boundary layers but measure different aspects of them: $\mathcal L_g\propto D^{-1}(\pi^2/6+\Delta)$ is dominated by the deterministic boundary-layer width $\propto1/D$, while $\epsilon_g\propto\sqrt{\Delta}/D$ is a noise-to-signal angle of the student against the teacher. The factor $D^{-2}$ in $\dot D$ connects to the derivative of the \mbox{scale-$D^{-1}$} loss, while the $\eta^2\mathcal{B}(\Delta)/b$ term in $\dot\Delta$ is the surviving online-SGD noise. Fixed-learning-rate dynamics therefore does not drive $\Delta$ to zero; it drives $\Delta$ to a noise floor while $D$ continues to grow:

\begin{corollary}[Fixed learning rate: one-third law]\label[corollary]{cor:fixed}
For constant $\eta>0$ and fixed $b$, \cref{eq:Deltadot-asymptotic-main} has the fixed point $2\Delta_* = \frac{\eta}{b}\,\mathcal{B}(\Delta_*)$
and the limiting trajectory satisfies
\begin{equation}
    \Delta(\alpha)\to\Delta_*\ ,
    \qquad
    D(\alpha)
    \sim
    \left[\frac{3Kc_K}{2}\,\eta\left(\frac{\pi^2}{6}+\Delta_*\right)\right]^{1/3}\!\alpha^{1/3}\ .
    \label{eq:fixedeta-centered-laws-main}
\end{equation}
Consequently, $\mathcal L_g(\alpha)\propto\alpha^{-1/3}$ and $\epsilon_g(\alpha)\propto\alpha^{-1/3}$.
\end{corollary}
The derivation is given in \cref{app:fixed-rate-derivation}. Within this solvable model, the one-third exponent is therefore not a consequence of a data spectrum. It follows from the boundary-layer drift $\dot D\propto D^{-2}$ together with a finite residual online-noise scale. Setting the online-noise term $\eta^2\mathcal B(\Delta)/b$ in \cref{eq:Deltadot-asymptotic-main} to zero (population gradient flow) would let $\Delta$ vanish faster than any power while $D\propto\alpha^{1/3}$ persists: the loss keeps its $\alpha^{-1/3}$ decay, consistent with \citet{liu2026universal}, while the error $\epsilon_g\propto\sqrt{\Delta}/D$ decays faster than any power. The one-third loss law is deterministic, the one-third error law is an online-SGD effect, invisible to gradient-flow analyses. A direct decomposition confirming that alignment growth and errors localize in the predicted boundary layers is reported in \cref{app:boundary-decomposition}.

\begin{figure}[t]
  \centering
  \includegraphics[]{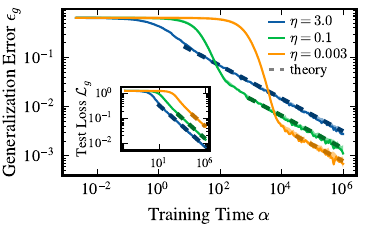}
  \includegraphics[]{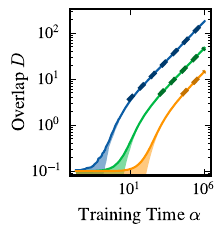}
  \includegraphics[]{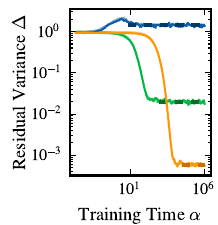}
  \caption{Finite-$N$ validation for fixed learning rates in the \(K=3\) online teacher--student model. The panels show the generalization error, centered overlap \(D\), and residual variance \(\Delta\) as functions of macroscopic time \(\alpha=\mu/N\). The curves show representative seed trajectories, with envelopes indicating fluctuations across six simulation seeds. Within these fluctuations, the trajectories agree with the predicted power-law prefactors and exponents: \(D\propto \alpha^{1/3}\), \(\Delta\) approaches a learning-rate-dependent floor, and
  \(\epsilon_g\propto \sqrt{\Delta}/D\propto \alpha^{-1/3}\); see \cref{eq:fixedeta-centered-laws-main,eq:epsg-boundary-main}.}
  \label{fig:fixed-eta-validation}
\end{figure}

\section{Exponents with learning rate schedules}
\label{sec:annealing}

Taking a smaller fixed learning rate lowers the asymptotic noise floor, but it also delays entry into the asymptotic regime; see \cref{fig:fixed-eta-validation}. Learning-rate schedules can improve the asymptotic error exponent because they allow the residual variance $\Delta$ to decrease while the accumulated learning time continues to grow. Suppose $\eta(\alpha)\to 0$ slowly enough that $\Delta$ adiabatically tracks the instantaneous fixed point $\Delta_*(\eta(\alpha))$ of \cref{eq:Deltadot-asymptotic-main}. Adiabatic means that the relaxation rate $\lambda\simeq 2Kc_K\eta/D$ toward this fixed point dominates the schedule's relative rate of change $|\eta'|/\eta$. A small-$\Delta$ expansion of $\mathcal{B}$ gives
$    \Delta(\alpha)
    \sim
    \frac{\kappa}{b}\,\eta(\alpha)$ with
$    \kappa=(2\log 2-1)/2$.
Then \cref{eq:Ddot-asymptotic-main} reduces to
%\begin{equation}
    $\dot D
    \sim
    \frac{Kc_K\pi^2}{12}\frac{\eta(\alpha)}{D^2}
    \label{eq:Ddot-schedule-leading-main}$ .
%\end{equation}
With
\begin{equation}
    H(\alpha)=\int_0^\alpha \eta(\alpha')\,d\alpha',
    \label{eq:H-main}
\end{equation}
one obtains
\begin{equation}
    D(\alpha)\sim \left(\frac{Kc_K\pi^2}{4}H(\alpha)\right)^{1/3} \ , \qquad \Delta \sim \frac{\kappa}{b}\,\eta(\alpha) \ .
    \label{eq:D-Delta-schedule-leading-main}
\end{equation}
Together with \cref{eq:epsg-boundary-main}, this establishes the results for power-law schedules:

\begin{corollary}[Power-law schedules]\label[corollary]{cor:sched}
For $\eta(\alpha)=\eta_0(1+\alpha/\alpha_0)^{-\gamma}$ with $0<\gamma<1$, the adiabatic tracking criterion holds, \cref{eq:D-Delta-schedule-leading-main} applies, and
\begin{equation}
    \epsilon_g(\alpha)
    \sim
    A_{K,b}\,\frac{\sqrt{\eta(\alpha)}}{H(\alpha)^{1/3}}
    \propto\alpha^{-(2+\gamma)/6} \ ,
    \qquad
    A_{K,b}=\Gamma_K\sqrt{\frac{\kappa}{b}}\left(\frac{4}{Kc_K\pi^2}\right)^{1/3} \ ,
    \label{eq:general-schedule-law-main}
\end{equation}
while by Theorem~\ref{thm:blexp} the loss slows to $\mathcal L_g\propto\alpha^{-(1-\gamma)/3}$.
\end{corollary}

The error exponent therefore interpolates continuously from $1/3$ at $\gamma=0$ toward $1/2$ as $\gamma \uparrow 1$. The endpoint is genuinely singular:

\begin{proposition}[Borderline schedule]\label{prop:borderline}
For $\gamma=1$, $\Delta$ does not generically track the instantaneous fixed point: $\Delta(\alpha)/\eta(\alpha)\nrightarrow\kappa/b$. Hence Corollary~\ref{cor:sched} does not apply at this endpoint.
\end{proposition}

\begin{figure}[t]
  \centering
  \includegraphics[]{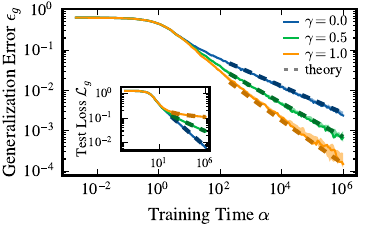}
  \includegraphics[]{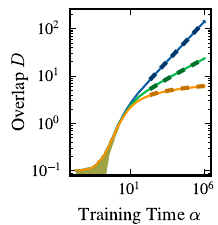}
  \includegraphics[]{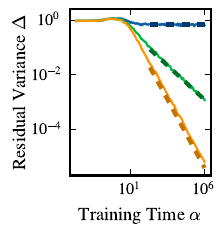}
  \caption{Schedule dependence in the \(K=3\) online teacher--student model. For
  \(\eta(\alpha)\propto \alpha^{-\gamma}\), the theory predicts
  \(\epsilon_g(\alpha)\propto \alpha^{-(2+\gamma)/6}\) for \(0\leq\gamma<1\). Increasing \(\gamma\) slows the growth of the
  centered overlap, \(D\propto \alpha^{(1-\gamma)/3}\) for \(\gamma<1\), but
  decreases the residual variance, \(\Delta\propto \eta(\alpha)\); the latter
  effect improves the classification-error exponent. The \(\gamma=1\) curve is a borderline case, where the adiabatic approximation for $\Delta$ breaks down. For reference, the adiabatic predictions of \cref{eq:D-Delta-schedule-leading-main,eq:general-schedule-law-main} are also shown.}
  \label{fig:schedule-validation}
\end{figure}

Thus the exponent $1/2$ is approached from below within the adiabatic family, but is not attained by the borderline schedule itself (see \cref{app:borderline-schedule}). If $\gamma>1$, the integral $H(\alpha)$ converges and the centered overlap saturates at a finite value, so both the test loss and the classification error approach nonzero constants and power-law decay ceases.  The derivations are collected in \cref{app:schedule-derivation}.

\section{Numerical validation and controlled departures}
\label{sec:numerics}

The simulations test the mechanism, not a universal empirical scaling law, first within the solvable model and then with correlated Gaussian inputs and whitened pretrained features as controlled departures from the theoretical assumptions. Numerical details, additional fixed-learning-rate sweeps, and the real-label whitened-feature comparison are reported in \cref{app:numerical-details}.

At fixed learning rate, the finite-\(N\) simulations in \cref{fig:fixed-eta-validation} agree with Theorem~\ref{thm:blexp} and Corollary~\ref{cor:fixed}, including test-loss decay, larger residual floors at larger learning rates, and later asymptotic onset at smaller ones. Additional simulations confirm the predicted \(K\)-dependence (\cref{app:K-dependence}), batch-size dependence for $b>1$ (\cref{app:b-dependence}), and increasing localization of active examples and errors in the predicted boundary layers as $D$ grows (\cref{app:boundary-decomposition}). The schedule sweep (\cref{fig:schedule-validation}) confirms both components of \cref{eq:D-Delta-schedule-leading-main}: annealing decreases \(\Delta\) but slows the growth of accumulated learning time \(H(\alpha)\) and \(D\), improving the error exponent as in \cref{eq:general-schedule-law-main} while slowing loss decay. For \(\gamma=1\), adiabatic tracking fails asymptotically, yet the adiabatic prediction remains a good finite-time reference over the displayed range, where the error shows a near-\(1/2\) power law.

\begin{figure}[t]
  \centering
  \includegraphics[]{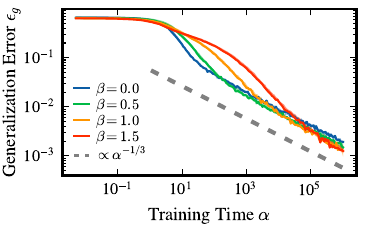}
  \hspace{1.5em}
  \includegraphics[]{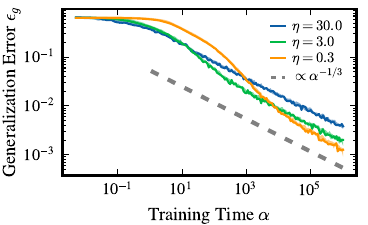}
  %\addtocounter{figure}{-1}
   \caption{Controlled departure from isotropic inputs.  Inputs are Gaussian with
  diagonal covariance spectrum \(\langle \xi_i^2\rangle\propto i^{-\beta}\),
  normalized so that the largest variance is one.  Left: \(\eta=0.5\) is fixed
  and \(\beta\) is varied.  Right: \(\beta=1.0\) is fixed and the learning rate is
  varied.  Increasing \(\beta\) changes and lengthens the transient regime.  The
  late-time decay remains consistent with the same boundary-layer asymptote, while
  smaller learning rates reduce asymptotic prefactors but can delay entry into the
  asymptotic regime.}
  \label{fig:correlated-gaussian}
\end{figure}

The correlated-Gaussian experiment (\cref{fig:correlated-gaussian}) suggests that structured covariance mainly changes the crossover: larger \(\beta\) lengthens transients, while later error decay remains consistent with the boundary-layer prediction. This supports the interpretation that covariance structure and boundary-layer dynamics can control different time windows.

\cref{fig:vit-whitened} relaxes the Gaussian assumption while controlling the covariance. The teacher-label run is consistent with the same qualitative scaling picture, suggesting that the mechanism is not immediately destroyed by non-Gaussian feature statistics. The corresponding real-label run (\cref{fig:app-vit-comparison} in \cref{app:vit_features}) reaches a floor much earlier. We interpret this as a limitation of the model. For real labels, aspects such as model misspecification, label noise, or irreducible Bayes error can dominate the late-time error before the noiseless boundary-layer asymptote is cleanly visible.

\cref{app:mnist-pythia} adds two practical comparisons. Batch-size-one SGD on MNIST is consistent with one-third scaling of both test loss and generalization error. Further, we analyze the Pythia Transformer family \citep{biderman2023pythia} similar to \citet{liu2026universal}, but concentrate on token top-1 error instead of loss. 
{Token top-1 error probes relative-logit dynamics. Its excess above a fitted floor is well described by a power law over the selected window, with an exponent compatible with one-third for the $12$-billion-parameter network.}

\section{Discussion and scope}
\label{sec:discussion}

\begin{wrapfigure}{R}{0.45\textwidth}
  \vspace{-0.8em}
  \centering
  \includegraphics{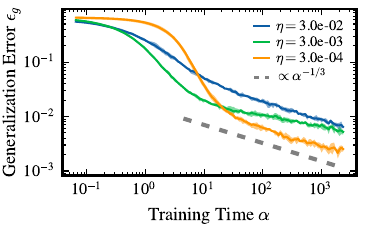}
  %\addtocounter{figure}{1}
    \caption{Non-Gaussian robustness test using whitened pretrained features. A
  linear softmax readout is trained online on whitened pretrained vision-transformer~(ViT) features \citep{dosovitskiy2021vit} of CIFAR-5M images \citep{nakkiran2021deep} with
  teacher-generated labels. Whitening controls the covariance, while the feature
  distribution remains non-Gaussian. The teacher-generated labels avoid an early
  real-label performance floor and allow the late-time regime to be observed.}
  \label{fig:vit-whitened}
  \vspace{-0.8em}
\end{wrapfigure}
\stepcounter{figure}

We have identified a boundary-layer mechanism for late-time online hard-label classification trained with softmax and cross-entropy in a solvable fixed-feature teacher-student model. For fixed learning rate, online gradient noise keeps the residual variance $\Delta$ finite, while only $O(D^{-1})$-thin decision-boundary layers remain active, which produces $D\propto \alpha^{1/3}$ and $\epsilon_g\propto \alpha^{-1/3}$.

Real-world learning problems can involve sub-tasks that occur at very different frequencies \citep{michaud2023quantization}.
\Cref{app:zipf} aggregates the fixed-rate law over Zipf-distributed sub-tasks (frequency $\propto k^{-\nu}$) in a minimal routed-expert model. The total error decays as $\alpha^{-\min\{1/3,\,c_\nu\}}$ with $c_\nu=1-1/\nu$, so heavy frequency tails can mask the $1/3$ exponent in aggregate observables, whereas a student that represents only the most frequent sub-tasks approaches its coverage floor as $\alpha^{-1/3}$, and frequency-resolved errors show the $1/3$ for every sub-task. This motivates the power-law-plus-floor fits in \Cref{app:mnist-pythia}: Pythia's extensive training at fixed model size could predominantly refine already represented patterns, as in the late-time regime of the Zipf model.

Our analytical results are derived under assumptions (A1)--(A7): a one-layer readout with fixed features, online SGD on fresh examples, the thermodynamic limit, noiseless hard labels, and permutation symmetry.  The theory does not claim that data structure is irrelevant, nor that feature learning cannot change the asymptotic class. Indeed, the correlated-input experiments show that structured covariance can indeed affect transients.

These limitations suggest concrete extensions. A soft-teacher or noisy-label version of the centered closure would connect the theory to Bayes-error floors in real data. Momentum should modify the online-noise term. Richer optimizers may change the range of useful learning-rate schedules, and feature-learning architectures could couple the boundary-layer dynamics of the readout to changes in the representation. The present work provides a baseline for those questions: a solvable classification scaling law whose exponent is generated by decision-boundary geometry and stochastic optimization noise.

\subsection*{Acknowledgments}
M.K. acknowledges financial support by the Federal Ministry of Research, Technology and Space of Germany and by the Sächsische Staatsministerium für Wissenschaft, Kultur und Tourismus in the programme Center of Excellence for AI-research ``Center for Scalable Data Analytics and Artificial Intelligence Dresden/Leipzig'', project identification number: ScaDS.AI.
The work of B.R.~is supported in part by the Simons Foundation Grant No.~12574. 
This work benefitted from discussions at Aspen Center for Physics, which is supported by National Science Foundation grant PHY-2210452.

\subsection*{AI use statement}

In this work, we used generative AI tools for the following tasks with required disclosure: help develop theoretical models, formulate mathematical claims, provide critical ingredients for proving mathematical claims, assist in the writing of proofs, provide feedback on research methodology and experiments, implement methods.
We have not used generative AI tools for the following tasks: propose or refine hypotheses, design research methodology or experiments, assist with translation, support qualitative and thematic data analysis, interpret results. 
The following tasks are not applicable to this work: generate synthetic data sets, clean and reformat dataset.
Additionally, we used generative AI tools for the following tasks with recommended
disclosure: create or edit software code, draft parts of a research paper, summarize or analyse existing literature, searching for information, edit a research paper to improve readability, identify relevant literature, propose a title or keywords for a research paper. 
We have reviewed all AI-assisted work. We take responsibility for the final content of this work, including text, claims or artifacts produced with the aid of generative AI.

\subsection*{Reproducibility statement}

We are committed to the reproducibility of our research. Our theoretical claims, including all assumptions, are presented in \cref{sec:centered-dynamics,sec:boundary-layers,sec:annealing} with derivations provided in \cref{app:theory}. Comprehensive details for reproducing our empirical results, including model architectures, data processing, and hyperparameter settings, are well documented in \cref{app:numerical-details}.
Source code is available under \cite{paperrepo}.

\bibliography{class_scalinglaw}

@article{richert2022soft,
  title={{S}oft mode in the dynamics of over-realizable online learning for soft committee machines},
  author={Richert, Frederieke and Worschech, Roman and Rosenow, Bernd},
  journal={Physical Review E},
  volume={105},
  number={5},
  pages={L052302},
  year={2022},
  publisher={APS}
}

@inproceedings{barkeshli2026origin,
  title={{O}n the {O}rigin of {N}eural {S}caling {L}aws: from {R}andom {G}raphs to {N}atural {L}anguage},
author={Maissam Barkeshli and Alberto Alfarano and Andrey Gromov},
booktitle={Forty-third International Conference on Machine Learning},
year={2026},
}

@article{cornacchia2023learning,
  title = {{L}earning {C}urves for the {M}ulti-{C}lass {T}eacher--{S}tudent {P}erceptron},
  author = {Cornacchia, Elisabetta and Mignacco, Francesca and Veiga, Rodrigo and Gerbelot, C{\'e}dric and Loureiro, Bruno and Zdeborov{\'a}, Lenka},
  journal = {Machine Learning: Science and Technology},
  volume = {4},
  number = {1},
  pages = {015019},
  year = {2023},
  publisher = {IOP Publishing},
}

@article{bahri2024explaining,
  title={{E}xplaining neural scaling laws},
  author={Bahri, Yasaman and Dyer, Ethan and Kaplan, Jared and Lee, Jaehoon and Sharma, Utkarsh},
  journal={Proceedings of the National Academy of Sciences},
  volume={121},
  number={27},
  pages={e2311878121},
  year={2024},
  publisher={National Academy of Sciences}
}

@article{sharma2022scaling,
  title={{S}caling laws from the data manifold dimension},
  author={Sharma, Utkarsh and Kaplan, Jared},
  journal={Journal of Machine Learning Research},
  volume={23},
  number={9},
  pages={1--34},
  year={2022}
}

@article{maloney2022solvable,
  title={{A} solvable model of neural scaling laws},
  author={Maloney, Alexander and Roberts, Daniel A and Sully, James},
  journal={arXiv preprint arXiv:2210.16859},
  year={2022}
}

@inproceedings{bordelon2020spectrum,
  title={{S}pectrum dependent learning curves in kernel regression and wide neural networks},
  author={Bordelon, Blake and Canatar, Abdulkadir and Pehlevan, Cengiz},
  booktitle={International Conference on Machine Learning},
  pages={1024--1034},
  year={2020},
  organization={PMLR}
}

@article{canatar2021spectral,
  title={{S}pectral bias and task-model alignment explain generalization in kernel regression and infinitely wide neural networks},
  author={Canatar, Abdulkadir and Bordelon, Blake and Pehlevan, Cengiz},
  journal={Nature communications},
  volume={12},
  number={1},
  pages={2914},
  year={2021},
  publisher={Nature Publishing Group UK London}
}

@article{lin2024scalinglinear,
  title={{S}caling laws in linear regression: {C}ompute, parameters, and data},
  author={Lin, Licong and Wu, Jingfeng and Kakade, Sham M and Bartlett, Peter L and Lee, Jason D},
  journal={Advances in Neural Information Processing Systems},
  volume={37},
  pages={60556--60606},
  year={2024}
}

@inproceedings{worschech2025analyzing,
  title={{A}nalyzing neural scaling laws in two-layer networks with power-law data spectra},
  author={Worschech, Roman and Rosenow, Bernd},
  booktitle={International Conference on Learning Representations},
  volume={2025},
  pages={89098--89145},
  year={2025}
}

@article{bordelon2025featurelearning,
  year = {2025},
publisher = {IOP Publishing},
volume = {2025},
number = {8},
pages = {084002},
author = {Bordelon, Blake and Atanasov, Alexander and Pehlevan, Cengiz},
title = {{H}ow feature learning can improve neural scaling laws},
journal = {Journal of Statistical Mechanics: Theory and Experiment},
}

@article{hestness2017deep,
  title={{D}eep learning scaling is predictable, empirically},
  author={Hestness, Joel and Narang, Sharan and Ardalani, Newsha and Diamos, Gregory and Jun, Heewoo and Kianinejad, Hassan and Patwary, Md Mostofa Ali and Yang, Yang and Zhou, Yanqi},
  journal={arXiv preprint arXiv:1712.00409},
  year={2017}
}

@article{kaplan2020scaling,
  title={{S}caling laws for neural language models},
  author={Kaplan, Jared and McCandlish, Sam and Henighan, Tom and Brown, Tom B and Chess, Benjamin and Child, Rewon and Gray, Scott and Radford, Alec and Wu, Jeffrey and Amodei, Dario},
  journal={arXiv preprint arXiv:2001.08361},
  year={2020}
}

@article{hoffmann2022chinchilla,
  title={{A}n empirical analysis of compute-optimal large language model training},
  author={Hoffmann, Jordan and Borgeaud, Sebastian and Mensch, Arthur and Buchatskaya, Elena and Cai, Trevor and Rutherford, Eliza and de Las Casas, Diego and Hendricks, Lisa Anne and Welbl, Johannes and Clark, Aidan and others},
  journal={Advances in neural information processing systems},
  volume={35},
  pages={30016--30030},
  year={2022}
}

@article{saad1995exact,
  title={{E}xact solution for on-line learning in multilayer neural networks},
  author={Saad, David and Solla, Sara A},
  journal={Physical Review Letters},
  volume={74},
  number={21},
  pages={4337},
  year={1995},
  publisher={APS}
}

@article{goldt2019dynamics,
  title={{D}ynamics of stochastic gradient descent for two-layer neural networks in the teacher-student setup},
  author={Goldt, Sebastian and Advani, Madhu and Saxe, Andrew M and Krzakala, Florent and Zdeborov{\'a}, Lenka},
  journal={Advances in neural information processing systems},
  volume={32},
  year={2019}
}

@article{advani2020high,
  title={{H}igh-dimensional dynamics of generalization error in neural networks},
  author={Advani, Madhu S and Saxe, Andrew M and Sompolinsky, Haim},
  journal={Neural Networks},
  volume={132},
  pages={428--446},
  year={2020},
  publisher={Elsevier}
}

@inproceedings{
bordelon2024dynamical,
title={{A} {D}ynamical {M}odel of {N}eural {S}caling {L}aws},
author={Blake Bordelon and Alexander Atanasov and Cengiz Pehlevan},
booktitle={Forty-first International Conference on Machine Learning},
year={2024},
}

@inproceedings{
liu2026universal,
title={{U}niversal {O}ne-third {T}ime {S}caling in {L}earning {P}eaked {D}istributions},
author={Yizhou Liu and Ziming Liu and Cengiz Pehlevan and Jeff Gore},
booktitle={Forty-third International Conference on Machine Learning},
year={2026},
}

@article{soudry2018implicit,
  title={{T}he implicit bias of gradient descent on separable data},
  author={Soudry, Daniel and Hoffer, Elad and Nacson, Mor Shpigel and Gunasekar, Suriya and Srebro, Nathan},
  journal={Journal of Machine Learning Research},
  volume={19},
  number={70},
  pages={1--57},
  year={2018}
}

@article{mignacco2020dynamical,
  title={{D}ynamical mean-field theory for stochastic gradient descent in gaussian mixture classification},
  author={Mignacco, Francesca and Krzakala, Florent and Urbani, Pierfrancesco and Zdeborov{\'a}, Lenka},
  journal={Advances in Neural Information Processing Systems},
  volume={33},
  pages={9540--9550},
  year={2020}
}

@article{loureiro2021learning,
  title={{L}earning curves of generic features maps for realistic datasets with a teacher-student model},
  author={Loureiro, Bruno and Gerbelot, Cedric and Cui, Hugo and Goldt, Sebastian and Krzakala, Florent and Mezard, Marc and Zdeborov{\'a}, Lenka},
  journal={Advances in Neural Information Processing Systems},
  volume={34},
  pages={18137--18151},
  year={2021}
}

@article{raissi2019physics,
  title={{P}hysics-informed neural networks: {A} deep learning framework for solving forward and inverse problems involving nonlinear partial differential equations},
  author={Raissi, Maziar and Perdikaris, Paris and Karniadakis, George E},
  journal={Journal of Computational physics},
  volume={378},
  pages={686--707},
  year={2019},
  publisher={Elsevier}
}

@article{nakada2020adaptive,
  title={{A}daptive approximation and generalization of deep neural network with intrinsic dimensionality},
  author={Nakada, Ryumei and Imaizumi, Masaaki},
  journal={Journal of Machine Learning Research},
  volume={21},
  number={174},
  pages={1--38},
  year={2020}
}

@inproceedings{opper1991calculation,
  title={{C}alculation of the learning curve of {B}ayes optimal classification algorithm for learning a perceptron with noise},
  author={Opper, Manfred and Haussler, David},
  booktitle={Colt},
  volume={91},
  pages={75--87},
  year={1991}
}

@article{biehl1994online,
  title={{O}n-{L}ine {L}earning with a {P}erceptron},
  author={Biehl, Michael and Riegler, Peter},
  journal={Europhysics Letters},
  volume={28},
  number={7},
  pages={525--530},
  year={1994},
  publisher={IOP Publishing}
}

@article{aubin2020perceptron,
  title={{G}eneralization error in high-dimensional perceptrons: {A}pproaching bayes error with convex optimization},
  author={Aubin, Benjamin and Krzakala, Florent and Lu, Yue and Zdeborov{\'a}, Lenka},
  journal={Advances in Neural Information Processing Systems},
  volume={33},
  pages={12199--12210},
  year={2020}
}

@inproceedings{
dosovitskiy2021vit,
title={{A}n {I}mage is {W}orth 16x16 {W}ords: {T}ransformers for {I}mage {R}ecognition at {S}cale},
author={Alexey Dosovitskiy and Lucas Beyer and Alexander Kolesnikov and Dirk Weissenborn and Xiaohua Zhai and Thomas Unterthiner and Mostafa Dehghani and Matthias Minderer and Georg Heigold and Sylvain Gelly and Jakob Uszkoreit and Neil Houlsby},
booktitle={International Conference on Learning Representations},
year={2021},
}

@article{mao2024surrogate,
  title={{A} universal growth rate for learning with smooth surrogate losses},
  author={Mao, Anqi and Mohri, Mehryar and Zhong, Yutao},
  journal={Advances in neural information processing systems},
  volume={37},
  pages={41670--41708},
  year={2024}
}

@article{rae2021scaling,
  title={{S}caling language models: {M}ethods, analysis \& insights from training gopher},
  author={Rae, Jack W and Borgeaud, Sebastian and Cai, Trevor and Millican, Katie and Hoffmann, Jordan and Song, Francis and Aslanides, John and Henderson, Sarah and Ring, Roman and Young, Susannah and others},
  journal={arXiv preprint arXiv:2112.11446},
  year={2021}
}

@article{zhang2004statistical,
  title={{S}tatistical behavior and consistency of classification methods based on convex risk minimization},
  author={Zhang, Tong},
  journal={The Annals of Statistics},
  volume={32},
  number={1},
  pages={56--85},
  year={2004},
  publisher={Institute of Mathematical Statistics}
}

@article{bartlett2006convexity,
  title={{C}onvexity, classification, and risk bounds},
  author={Bartlett, Peter L and Jordan, Michael I and McAuliffe, Jon D},
  journal={Journal of the American Statistical Association},
  volume={101},
  number={473},
  pages={138--156},
  year={2006},
  publisher={Taylor \& Francis}
}

@inproceedings{nacson2019stochastic,
  title={{S}tochastic gradient descent on separable data: {E}xact convergence with a fixed learning rate},
  author={Nacson, Mor Shpigel and Srebro, Nathan and Soudry, Daniel},
  booktitle={The 22nd International Conference on Artificial Intelligence and Statistics},
  pages={3051--3059},
  year={2019},
  organization={PMLR}
}

@article{wang2024unified,
  title={{U}nified binary and multiclass margin-based classification},
  author={Wang, Yutong and Scott, Clayton},
  journal={Journal of Machine Learning Research},
  volume={25},
  number={143},
  pages={1--51},
  year={2024}
}

@article{ravi2024implicit,
  title={{T}he implicit bias of gradient descent on separable multiclass data},
  author={Ravi, Hrithik and Scott, Clayton and Soudry, Daniel and Wang, Yutong},
  journal={Advances in neural information processing systems},
  volume={37},
  pages={81324--81359},
  year={2024}
}

@article{seung1992statistical,
  title={{S}tatistical mechanics of learning from examples},
  author={Seung, Hyunjune Sebastian and Sompolinsky, Haim and Tishby, Naftali},
  journal={Physical review A},
  volume={45},
  number={8},
  pages={6056},
  year={1992},
  publisher={APS}
}

@inproceedings{
nakkiran2021deep,
title={{T}he {D}eep {B}ootstrap {F}ramework: {G}ood {O}nline {L}earners are {G}ood {O}ffline {G}eneralizers},
author={Preetum Nakkiran and Behnam Neyshabur and Hanie Sedghi},
booktitle={International Conference on Learning Representations},
year={2021},
}

@inproceedings{gadre2025language,
  title={{L}anguage models scale reliably with over-training and on downstream tasks},
  author={Gadre, Samir Yitzhak and Smyrnis, Georgios and Shankar, Vaishaal and Gururangan, Suchin and Wortsman, Mitchell and Shao, Rulin and Mercat, Jean and Fang, Alex and Li, Jeffrey and Keh, Sedrick and others},
  booktitle={International Conference on Learning Representations},
  volume={2025},
  pages={67661--67682},
  year={2025}
}

@inproceedings{
schaeffer2025why,
title={{W}hy {H}as {P}redicting {D}ownstream {C}apabilities of {F}rontier {{A}{I}} {M}odels with {S}cale {R}emained {E}lusive?},
author={Rylan Schaeffer and Hailey Schoelkopf and Brando Miranda and Gabriel Mukobi and Varun Madan and Adam Ibrahim and Herbie Bradley and Stella Biderman and Sanmi Koyejo},
booktitle={Forty-second International Conference on Machine Learning},
year={2025},
}

@inproceedings{yan2026scales,
  title={{W}hat scales in cross-entropy scaling law?},
  author={Yan, Junxi and Wei, Zixi and Ai, Qingyao and Liu, Yiqun and Zhan, Jingtao},
  booktitle={International Conference on Learning Representations},
  volume={2026},
  pages={79741--79758},
  year={2026}
}

@inproceedings{biderman2023pythia,
  title={{P}ythia: {A} suite for analyzing large language models across training and scaling},
  author={Biderman, Stella and Schoelkopf, Hailey and Anthony, Quentin Gregory and Bradley, Herbie and O’Brien, Kyle and Hallahan, Eric and Khan, Mohammad Aflah and Purohit, Shivanshu and Prashanth, USVSN Sai and Raff, Edward and others},
  booktitle={International conference on machine learning},
  pages={2397--2430},
  year={2023},
  organization={PMLR}
}

@article{paccolat2021geometric,
  title={{G}eometric compression of invariant manifolds in neural networks},
  author={Paccolat, Jonas and Petrini, Leonardo and Geiger, Mario and Tyloo, Kevin and Wyart, Matthieu},
  journal={Journal of Statistical Mechanics: Theory and Experiment},
  volume={2021},
  number={4},
  pages={044001},
  year={2021},
  publisher={IOP Publishing and SISSA}
}

@article{sclocchi2024different,
  title={{O}n the different regimes of stochastic gradient descent},
  author={Sclocchi, Antonio and Wyart, Matthieu},
  journal={Proceedings of the National Academy of Sciences},
  volume={121},
  number={9},
  pages={e2316301121},
  year={2024},
  publisher={National Academy of Sciences}
}

@article{penedo2024fineweb,
  title={{T}he {F}ineWeb {D}atasets: {D}ecanting the {W}eb for the {F}inest {T}ext {D}ata at {S}cale},
  author={Penedo, Guilherme and Kydl\'{\i}\v{c}ek, Hynek and allal, Loubna Ben and Lozhkov, Anton and Mitchell, Margaret and Raffel, Colin and Von Werra, Leandro and Wolf, Thomas},
  journal={Advances in Neural Information Processing Systems},
  volume={37},
  pages={30811--30849},
  year={2024}
}

@article{hutter2021learning,
  title={{L}earning curve theory},
  author={Hutter, Marcus},
  journal={arXiv preprint arXiv:2102.04074},
  year={2021}
}

@article{michaud2023quantization,
  title={{T}he {Q}uantization {M}odel of {N}eural {S}caling},
  author={Michaud, Eric and Liu, Ziming and Girit, Uzay and Tegmark, Max},
  journal={Advances in Neural Information Processing Systems},
  volume={36},
  pages={28699--28722},
  year={2023}
}

@article{nam2024exactly,
  title={An exactly solvable model for emergence and scaling laws in the multitask sparse parity problem},
  author={Nam, Yoonsoo and Fonseca, Nayara and Lee, Seok H and Mingard, Chris and Louis, Ard A},
  journal={Advances in Neural Information Processing Systems},
  volume={37},
  pages={39632--39693},
  year={2024}
}

@inproceedings{zhai2022scaling,
  title={{S}caling vision transformers},
  author={Zhai, Xiaohua and Kolesnikov, Alexander and Houlsby, Neil and Beyer, Lucas},
  booktitle={2022 IEEE/CVF Conference on Computer Vision and Pattern Recognition (CVPR)},
  pages={1204--1213},
  year={2022},
  organization={IEEE}
}

@inproceedings{deng2009imagenet,
  title={{I}magenet: {A} large-scale hierarchical image database},
  author={Deng, Jia and Dong, Wei and Socher, Richard and Li, Li-Jia and Li, Kai and Fei-Fei, Li},
  booktitle={2009 IEEE conference on computer vision and pattern recognition},
  pages={248--255},
  year={2009},
  organization={Ieee}
}

@misc{paperrepo,
    author={K\"uhn, Marcel and Thelge, Yoon and Rosenow, Bernd},
    year={2026},
    title={Github repository, \url{https://github.com/RosenowGroup/classification-one-third-scaling}},
}
\bibliographystyle{iclr2027_conference}

\appendix

\section{Proofs of the theoretical results}
\label{app:theory}

This section proves Lemmas~\ref{lem:closure} and~\ref{lem:entry}, Theorem~\ref{thm:blexp}, and Corollaries~\ref{cor:fixed} and~\ref{cor:sched} in order. Proposition~\ref{prop:borderline} is treated together with the last corollary. Throughout, $K\ge2$ and the fresh batch size $b$ are fixed. We first take the thermodynamic limit at fixed $\alpha=\mu/N$, as in (A6), and then study the late-time behavior of the resulting ODE.

\subsection{Lemma \ref{lem:closure}: centered closure}
\label{app:exact-K-dynamics}

The teacher fields $u_a=T_a\cdot\xi/\sqrt N$ are i.i.d.~standard Gaussians, and $t_a=J_a\cdot\xi/\sqrt N$ are the student logits. Under the permutation-symmetric ansatz, the macroscopic state is described by
\begin{equation}
R=\frac{J_1\cdot T_1}{N} \ ,\qquad
S=\frac{J_1\cdot T_2}{N} \ ,\qquad
Q=\frac{J_1\cdot J_1}{N} \ ,\qquad
C=\frac{J_1\cdot J_2}{N} \ .
\label{eq:app-RSQC}
\end{equation}
For a fresh batch of $b$ independent examples, the update is
\begin{equation}
J_a^{\mu+1}=J_a^\mu+\frac{\eta}{b\sqrt N}\sum_{i=1}^b g_a^i\xi^i \ ,
\qquad g_a^i=p_a^{{\rm T},i}-p_a^i \ .
\label{eq:app-softmax-update}
\end{equation}
Here $p_a^{\rm T}=\mathbf1\{a=y\}$, $y=\argmax_c u_c$, and $p_a=e^{t_a}/\sum_c e^{t_c}$. We use the standard thermodynamic-limit overlap equations of online teacher--student learning \citep{biehl1994online,saad1995exact}. Rigorous fixed-time convergence to such ODEs in closely related Gaussian teacher--student models is proven in \citet{goldt2019dynamics}. For a perceptron derivation retaining batch-dependent noise, see \citet{sclocchi2024different}. With our batch-averaged update and optimizer-step time, the corresponding overlap equations are
\begin{align}
\dot R&=\eta\langle g_1u_1\rangle \ ,\label{eq:app-Rdot}\\
\dot S&=\eta\langle g_1u_2\rangle \ ,\label{eq:app-Sdot}\\
\dot Q&=2\eta\langle g_1t_1\rangle+\frac{\eta^2}{b}\langle g_1^2\rangle \ ,\label{eq:app-Qdot}\\
\dot C&=\eta\langle g_1t_2+g_2t_1\rangle+\frac{\eta^2}{b}\langle g_1g_2\rangle \ .\label{eq:app-Cdot}
\end{align}
Brackets denote averages over the jointly Gaussian teacher and student fields at fixed order parameters. These are the thermodynamic-limit, $N\to \infty$, overlap equations, not connected to a late-time assumption. We do not repeat their stochastic-process convergence proof here.

For orientation, the single-step increments of $R$ and $Q$ are
\begin{align*}
\delta R&=\frac{\eta}{Nb}\sum_i g_1^i u_1^i \ ,\\
\delta Q&=\frac{2\eta}{Nb}\sum_i g_1^i t_1^i
 +\frac{\eta^2}{N^2b^2}\sum_{i,j}g_1^i g_1^j\,\xi^i\cdot\xi^j \ .
\end{align*}
Multiplication by $N$ converts these increments to derivatives in $\alpha$. In the quadratic term, the $b$ diagonal contributions have $\|\xi^i\|^2/N\to1$ and give the factor $1/b$. The cross-example terms have conditional mean $O(N^{-2})$ per update and vanish on this time scale. Thus batching changes the quadratic noise term, not the linear drift. Counting processed examples instead of optimizer steps would also rescale time, whereas we define it as $\alpha=\mu/N$ for all values of $b$.

\subsubsection{Centered Gaussian representation}
\label{app:centered-gaussian-representation}

Let $\bar u=K^{-1}\sum_a u_a$, $\bar t=K^{-1}\sum_a t_a$, and $h_a=t_a-\bar t$. Teacher labels and softmax probabilities are invariant under common shifts of their fields. A covariance calculation from \cref{eq:app-RSQC} gives
\begin{equation}
\langle h_a(u_c-\bar u)\rangle=D(\delta_{ac}-1/K) \ ,
\qquad
\langle h_ah_c\rangle=Q_{\mathrm{eff}}(\delta_{ac}-1/K) \ .
\end{equation}
Consequently, $\langle h_a\mid u_1,\ldots,u_K\rangle=D(u_a-\bar u)$. The residual is independent of the teacher fields, is Gaussian, and has covariance $\Delta(\delta_{ac}-1/K)$. This establishes $\Delta\ge0$ and the representation
\begin{equation}
h_a=D(u_a-\bar u)+\sqrt\Delta\,(z_a-\bar z) \ ,
\qquad \Delta=Q_{\mathrm{eff}}-D^2 \ ,
\label{eq:app-centered-logits}
\end{equation}
where the $z_a$ are independent standard Gaussians, independent of the $u_a$, and $\bar z=K^{-1}\sum_a z_a$.

\subsubsection{Exact centered closure}
\label{app:exact-centered-closure}

Subtracting \cref{eq:app-Sdot} from \cref{eq:app-Rdot} and using class symmetry gives
\begin{equation}
\dot D=\eta\langle g_1(u_1-u_2)\rangle
=\frac{K\eta}{K-1}\langle g_1(u_1-\bar u)\rangle \ .
\label{eq:app-Ddot-exact-centered}
\end{equation}
Likewise, $\sum_a g_a=0$ implies $\langle g_1g_2\rangle=-\langle g_1^2\rangle/(K-1)$. Subtracting \cref{eq:app-Cdot} from \cref{eq:app-Qdot} therefore yields
\begin{align}
\dot Q_{\mathrm{eff}}&=\frac{K}{K-1}
\left[2\eta\langle g_1h_1\rangle+\frac{\eta^2}{b}\langle g_1^2\rangle\right],
\label{eq:app-Qeffdot-exact-centered}
\end{align}
and by definition of $\Delta$
\begin{align}
\dot\Delta&=\dot Q_{\mathrm{eff}}-2D\dot D \ .
\label{eq:app-Deltadot-exact-centered}
\end{align}
This proves \cref{lem:closure}. Two equivalent forms will be useful below. Gaussian integration by parts in the independent $z_a$ gives
$\langle g_1(z_1-\bar z)\rangle=-\sqrt\Delta\langle p_1(1-p_1)\rangle$:
the common-mode derivative cancels because softmax is shift invariant, and $\partial_{h_1}g_1=-p_1(1-p_1)$. Thus
\begin{equation}
\dot\Delta=\frac{K\eta}{K-1}
\left[-2\Delta\langle p_1(1-p_1)\rangle+\frac{\eta}{b}\langle g_1^2\rangle\right].
\label{eq:app-Delta-source-contraction}
\end{equation}
The contraction is proportional to $\Delta$ exactly, before any asymptotic expansion. Also, summing $g_a u_a$ over classes gives
\begin{align}
\dot D&=\frac{\eta}{K-1}
\left\langle u_y-\sum_a p_a u_a\right\rangle \nonumber\\
&=\frac{\eta}{K-1}\left\langle\sum_a p_a(u_y-u_a)\right\rangle \nonumber \\
&\ge0 \ .
\label{eq:app-D-positive}
\end{align}

\subsection{Lemma \ref{lem:entry}: global entry into the late-time regime}
\label{app:global-entry}

The derivation first bounds the positive noise source term relative to the negative contraction term proportional to $\Delta$ in \cref{eq:app-Delta-source-contraction}, without assuming large $D$. This bounds $\Delta$; positivity of \cref{eq:app-D-positive} then forces $D$ to diverge.

\subsubsection{Comparing the source and contraction}

For $D\ge0$ and $\Delta\ge0$, we claim that
\begin{equation}
\langle g_1^2\rangle\le C_K(1+\sqrt\Delta)\langle p_1(1-p_1)\rangle 
\label{eq:app-source-comparison}
\end{equation}
for some constant $C_K$. To see this, use the uncentered logits $Du_a+\sqrt\Delta z_a$, which are independent Gaussians of variance $Q_{\mathrm{eff}}=D^2+\Delta$. Let $G$ denote the top gap of $K$ independent standard Gaussians. Choosing their ordered winner and runner-up gives its density
\begin{equation}
f_G(g)=K(K-1)\int_{-\infty}^\infty
\varphi(v+g)\varphi(v)\Phi(v)^{K-2}\,dv,\qquad g\ge0 \ .
\label{eq:app-top-gap-density}
\end{equation}
This density is bounded, is positive near zero, and has a bounded derivative. On $\sqrt{Q_{\mathrm{eff}}}G\le1$, the two largest softmax probabilities are at least $1/K$ and $e^{-1}/K$. Since $K\langle p_1(1-p_1)\rangle=\langle1-\sum_a p_a^2\rangle$, integration over this event gives a lower bound. Conversely, the mass outside the student winner $\hat y$ is at most $(K-1)e^{-\sqrt{Q_{\mathrm{eff}}}G}$. Integrating this bound gives
\begin{equation}
\langle p_1(1-p_1)\rangle\ge\frac{c}{1+\sqrt{Q_{\mathrm{eff}}}},
\qquad
\langle1-p_{\hat y}\rangle\le\frac{C}{1+\sqrt{Q_{\mathrm{eff}}}},
\label{eq:app-softmax-uncertainty-bounds}
\end{equation}
with positive constants depending only on $K$. These bounds include $\sqrt{Q_{\mathrm{eff}}}=0$, when softmax is uniform.

For $D>0$, a teacher--student disagreement requires the teacher top gap to be no larger than $(\sqrt\Delta/D)\max_{a,c}|z_a-z_c|$. The bounded gap density and finite Gaussian moments therefore imply
\begin{equation}
\Pr(\hat y\ne y)\le\min\{1,C_K\sqrt\Delta/D\}.
\end{equation}
Finally, $\sum_a g_a^2\le2(1-p_y)$ and
$1-p_y\le\mathbf1\{\hat y\ne y\}+1-p_{\hat y}$ give
\[
\frac{\langle g_1^2\rangle}{\langle p_1(1-p_1)\rangle}
\le C_K\bigl[1+(1+\sqrt{Q_{\mathrm{eff}}})\Pr(\hat y\ne y)\bigr]
\le C_K'(1+\sqrt\Delta).
\]
The last inequality follows by separating $D\le\max\{1,\sqrt\Delta\}$ from its complement; at $D=0$ use the trivial disagreement bound. This proves \cref{eq:app-source-comparison}.

\subsubsection{Bounded residual variance and diverging alignment}

Write $\mu_K=\langle\max_a u_a\rangle>0$. Integration by parts in $u_a$ gives the additional identity
\begin{equation}
\dot D=\frac{\eta}{K-1}
\left[\mu_K-KD\langle p_1(1-p_1)\rangle\right].
\label{eq:app-D-ibp}
\end{equation}
While $D\le0$, this is at least $\eta\mu_K/(K-1)$. Thus any negative initial alignment reaches zero after finite accumulated learning time. During this interval, \cref{eq:app-Delta-source-contraction} gives $\dot\Delta\le K\eta_{\max}\eta/[b(K-1)]$, so at the crossing
\[
\Delta\le\Delta(0)+\frac{K\eta_{\max}}{b\mu_K}(-D(0))_+.
\]
For $D\ge0$, \cref{eq:app-source-comparison} instead gives
\begin{equation}
\dot\Delta\le\frac{K\eta}{K-1}\langle p_1(1-p_1)\rangle
\left[-2\Delta+\frac{\eta}{b}C_K(1+\sqrt\Delta)\right].
\end{equation}
Choose $M$ above the initial/transient bound and large enough that
$2M>(\eta_{\max}/b)C_K(1+\sqrt M)$. At $\Delta=M$ the drift cannot point outward, so $0\le\Delta(\alpha)\le M$ for all times.

By \cref{eq:app-D-positive}, $D$ is nondecreasing. If it had a finite limit, $(D,\Delta)$ would remain in a compact set. At every finite state the average in \cref{eq:app-D-positive} is strictly positive: teacher ties have probability zero, and every softmax probability is positive. This average is continuous in $(D,\Delta)$, so it has a positive minimum on that compact set. Hence $\dot D\ge c\eta$, contradicting $\int_0^\alpha\eta\to\infty$ and boundedness of $D$. This proves \cref{lem:entry} without using the boundary-layer expansion.

\subsubsection{Heuristic late-time regime entry scale}
\label{app:heuristic_entry_scale}

Even though our theoretical analysis considers the limit $\alpha \to \infty$ and therefore $D \to \infty$, for finite-time numerics we can give a heuristic entry scale for $\alpha$ such that the boundary layer description becomes reasonable. Consider that the two largest Gaussian
teacher fields are typically separated by only \(O(1/\sqrt{2\log K})\), so the
corresponding student-logit gap is \(O(D/\sqrt{2\log K})\). Thus the classes are determined exponentially well away from decision boundaries under the condition \(D\gg\sqrt{2\log K}\). For bounded moderate residual variance, the crossing $D(\alpha_*)\approx\sqrt{2\log K}$ serves as a heuristic entry scale. Combining it with \cref{eq:fixedeta-centered-laws-main} and the leading large-$K$ estimate $c_K\approx\sqrt{2\log K}/K^2$ gives
\begin{equation}
\alpha_*\approx\frac{4K\log K}{3\eta\left(\pi^2/6+\Delta_*\right)},
\label{eq:heuritic_entry_scale}
\end{equation}
predicting later crossover for smaller $\eta$ and larger $K$.

\subsection{Theorem \ref{thm:blexp}: uniform boundary-layer expansion}
\label{app:boundary-layer-derivation}

\cref{lem:entry} places every trajectory, eventually, in the regime
\begin{equation}
D\to\infty \ ,
\qquad
0\le\Delta\le M<\infty \ .
\label{eq:app-boundary-regime}
\end{equation}
The task is now to evaluate the four Gaussian averages that the macroscopic dynamics needs, namely $\dot D$, the contraction $\langle p_1(1-p_1)\rangle$, the noise source $\langle g_1^2\rangle$ and the test loss $\mathcal L_g=\langle-\log p_y\rangle$, to leading order in this regime, with remainders that hold uniformly for $\Delta$ in $[0,M]$, so the expansion applies also when $\Delta$ varies with time or tends to zero.

We first consider geometry: how much input probability lies within a teacher boundary layer of width $O(D^{-1})$, the answer being $O(D^{-1})$ with a prefactor set by the boundary density $c_K$.  Then we consider the local problem: inside such a layer only two classes compete, and after rescaling the teacher gap by $D$ all four averages reduce to elementary one-dimensional integrals of explicit binary kernels. We show that everything else, meaning both the rest of the input space and the difference between the true multiclass integrand and the idealized binary one, contributes less.  That last part produces the remainders, and we identify which of them dominates.

\subsubsection{Boundary density and local binary integrals}
\label{app:boundary-density}
\label{app:local-integrals}

Fix an unordered pair $\{a,c\}$. On $u_a=u_c=s$, all other fields must lie below $s$ for this pair to compete for the label. The single-pair boundary density is therefore
\begin{equation}
c_K=\int_{-\infty}^\infty\varphi(s)^2\Phi(s)^{K-2}\,ds.
\label{eq:app-cK}
\end{equation}
For a nonnegative gap $q$, let $w_K(q)=f_G(q)/[K(K-1)]$ be the density for one specified ordered top pair. Then $w_K(0)=c_K$, and \cref{eq:app-top-gap-density} gives
\begin{equation}
|w_K(q)-c_K|\le C_Kq\quad(0\le q\le1),
\qquad
0\le w_K(q)\le\frac{e^{-q^2/4}}{2\sqrt\pi}\quad(q\ge0).
\label{eq:app-density-bounds}
\end{equation}
In particular, $\Pr(G\le\lambda/D)=K(K-1)c_K\lambda/D+O_{K,\lambda}(D^{-2})$ for fixed $\lambda$. This is the $O(D^{-1})$ measure of the active layer.

Near a boundary, write $u_a-u_c=x/D$. The student gap is then
$h_a-h_c=x+\delta$, where $\delta=\sqrt{2\Delta}\,z$ and $z\sim\mathcal N(0,1)$.
The local residual is $F_\delta(x):=\Theta(x)-\sigma(x+\delta)$, with $\sigma(y)=(1+e^{-y})^{-1}$. We use the elementary identities
\begin{align}
A_0(\delta)&:=\int_{\mathbb R}F_\delta(x)\,dx=-\delta,\label{eq:app-A0}\\
A_1(\delta)&:=\int_{\mathbb R}xF_\delta(x)\,dx=\frac{\pi^2}{6}+\frac{\delta^2}{2},\label{eq:app-A1}\\
A_2(\delta)&:=\int_{\mathbb R}(x+\delta)F_\delta(x)\,dx=\frac{\pi^2}{6}-\frac{\delta^2}{2},\label{eq:app-A2}\\
B_0(\delta)&:=\int_{\mathbb R}F_\delta(x)^2\,dx
=2\log\left(2\cosh\frac\delta2\right)-1.\label{eq:app-B0}
\end{align}
For example, differentiating $A_1$ gives $A_1'(\delta)=\delta$, while
$A_1(0)=2\int_0^\infty x/(1+e^x)\,dx=\pi^2/6$; also $A_2=A_1+\delta A_0$.
The local loss satisfies
\begin{equation}
\mathcal L_0(\delta):=\int_0^\infty
\bigl[\log(1+e^{-x-\delta})+\log(1+e^{-x+\delta})\bigr]\,dx
=\frac{\pi^2}{6}+\frac{\delta^2}{2}.
\label{eq:app-local-loss}
\end{equation}
Here differentiation twice in $\delta$ gives one, and the value at zero is $\pi^2/6$.
Averaging with $Dz=(2\pi)^{-1/2}e^{-z^2/2}\,dz$ yields
\begin{align}
\langle A_1(\delta)\rangle_\delta&=\frac{\pi^2}{6}+\Delta,\label{eq:app-A1-avg}\\
\langle A_2(\delta)\rangle_\delta&=\frac{\pi^2}{6}-\Delta,\label{eq:app-A2-avg}\\
\mathcal B(\Delta)&:=\langle B_0(\delta)\rangle_\delta
=\int Dz\left[2\log\left(2\cosh\left(\sqrt{\frac\Delta2}z\right)\right)-1\right].\label{eq:app-B-Delta}
\end{align}
In addition, symmetry of $\delta$ and $\int_{\mathbb R}\sigma'(x+\delta)\,dx=1$ give
$\int_0^\infty\langle\sigma'(x+\delta)\rangle_\delta\,dx=1/2$.

\subsubsection{Localization and the sources of the remainders}
\label{app:localization-remainders}

A binary calculation alone does not bound the multiclass error. We separate four regions before extending the binary integrals to the whole line. Let $I,J$ be the teacher winner and runner-up, $G=u_I-u_J$, and $G_2=u_{(2)}-u_{(3)}$; set $G_2=\infty$ for $K=2$. Choose $L=8\log D$ and first restrict to
\begin{equation}
\sqrt\Delta\max_{a,c}|z_a-z_c|\le L/4.
\label{eq:app-noise-cutoff}
\end{equation}
On this event, partition teacher configurations into the bulk $G>L/D$, an isolated pair $G\le L/D$, $G_2>4L/D$, and near-triple ties $G\le L/D$, $G_2\le4L/D$. 

\paragraph{Bulk and isolated pairs.}
In the bulk under \cref{eq:app-noise-cutoff}, every student gap from the teacher winner is at least $3L/4$. $p_1(1-p_1)$, $g_1^2$, and the loss are therefore $O_K(e^{-3L/4})$. For the alignment drift use \cref{eq:app-D-positive}: a competitor with teacher gap $t\ge G$ contributes at most
\[
t e^{-Dt+L/4}\le\frac{L}{D}e^{-3L/4},
\]
since $Dt\ge L>1$. Thus its bulk contribution is smaller still.

For an isolated pair, put $x=DG\in[0,L]$ and $\delta=\sqrt\Delta(z_I-z_J)$. Every other class is at least $3L$ below the runner-up in student-logit units, giving
\begin{equation}
p_I=\sigma(x+\delta)+O_K(e^{-3L}),\qquad
p_J=1-\sigma(x+\delta)+O_K(e^{-3L}).
\label{eq:app-isolated-softmax}
\end{equation}
In the binary limit $g_J=-g_I$, so the common boundary coordinate cancels in
$\sum_a g_a u_a=g_I(u_I-u_J)$. The alignment contribution is consequently
\begin{equation}
\frac{p_JG}{K-1}=\frac{1}{(K-1)D}\frac{x}{1+e^{x+\delta}}.
\label{eq:app-local-alignment}
\end{equation}
This exact cancellation before integration supplies one factor $D^{-1}$; the layer measure supplies the other. The corresponding kernels for contraction, source, and loss are $\sigma'(x+\delta)$, $(1-\sigma(x+\delta))^2$, and $\log(1+e^{-x-\delta})$. The exponentially small errors in \cref{eq:app-isolated-softmax}, including the teacher-gap-weighted third-class terms, are below the remainder orders stated below.

\paragraph{Near-triple ties.}
For $K\ge3$, the joint density of $(G,G_2)$ is bounded. For a specified ordered top triple it is obtained by integrating
$\varphi(s+g+h)\varphi(s+h)\varphi(s)\Phi(s)^{K-3}$ over $s$. Hence
\begin{equation}
\Pr(G\le L/D,\ G_2\le4L/D)=O_K(L^2/D^2).
\label{eq:app-near-triple-mass}
\end{equation}
The contraction and source integrands are bounded, giving $O_K(L^2/D^2)$ contributions. Under the noise bound the loss is at most $L+\log K$, giving $O_K(L^3/D^2)$. For alignment, each gap $t\le2L/D$ contributes at most $2L/D$; for $t>2L/D$, the bound $t e^{-Dt+L}\le(2L/D)e^{-L}$ applies. Thus the whole alignment integrand is $O_K(L/D)$, giving $O_K(L^3/D^3)$ after multiplication by the near-triple mass. The binary terms added across this set obey the same bounds. For $K=2$ this region is absent.

\paragraph{Noise and binary tails.}
For $0\le\Delta\le M$, a Gaussian union bound makes the probability of violating \cref{eq:app-noise-cutoff} at most $C_K e^{-c_M L^2}$ (zero when $M=0$). The loss is bounded by $\log K+\sqrt M\max_{a,c}|z_a-z_c|$, and the alignment integrand by the teacher-field range. Thus the relevant Gaussian moments are bounded uniformly in $D$, and Cauchy--Schwarz bounds the discarded contributions by a quantity smaller than every inverse power of $D$.
To remove the gap cutoff, consider any of the four nonnegative binary kernels
\[
k(x,\delta)\in\left\{\frac{x}{1+e^{x+\delta}},\ \sigma'(x+\delta),\
(1-\sigma(x+\delta))^2,\ \log(1+e^{-x-\delta})\right\}.
\]
Splitting the noise average into $\delta\ge-x/2$ and its complement gives, uniformly for $0\le\Delta\le M$,
\begin{equation}
\langle k(x,\delta)\rangle_\delta
\le C_M(1+x)\bigl(e^{-x/2}+e^{-c_Mx^2}\bigr),\qquad x\ge0.
\label{eq:app-kernel-tail}
\end{equation}
On the first event this follows from logistic tails; on the second, from the Gaussian tail and, for the loss, $\log(1+e^{-x-\delta})\le\log2+|\delta|$. In particular all polynomially weighted $x$ moments are uniformly finite. The global density bound in \cref{eq:app-density-bounds} then gives
\begin{equation}
\int_L^\infty w_K(x/D)\langle k(x,\delta)\rangle_\delta\,dx
\le C_{K,M}(1+L)\bigl(e^{-L/2}+e^{-c'_M L^2}\bigr).
\label{eq:app-cutoff-tail}
\end{equation}
With $L=8\log D$, this is $O_{K,M}((1+\log D)D^{-4})$ plus a superpolynomially small term. The outside prefactor is $D^{-2}$ for alignment and $D^{-1}$ for the other averages. Thus cutoff removal costs $o(D^{-3})$ and $o(D^{-2})$, respectively, and introduces no additional logarithm.

Finally, replacing $w_K(x/D)$ by $c_K$ costs $O_{K,M}(D^{-1})$ inside any of these integrals: for $x\le D$, use $|w_K(x/D)-c_K|\le C_Kx/D$ and the first weighted kernel moment; for $x>D$, use \cref{eq:app-kernel-tail}. The resulting density corrections are $O(D^{-3})$ for alignment and $O(D^{-2})$ otherwise. For $K\ge3$, the largest retained bounds are therefore the near-triple bounds, not the bulk, noise, density correction, or added tails. The logarithms reflect the chosen localization estimates; we do not claim that these bounds are sharp asymptotic corrections.

\subsubsection{Asymptotic flow and test loss}
\label{app:asymptotic-flow}

There are $K(K-1)$ ordered top pairs, each with gap density $w_K(q)\,dq$. Using \cref{eq:app-local-alignment} and $q=x/D$ therefore gives
\begin{align}
\dot D&=\frac{K\eta}{D^2}\int_0^\infty w_K(x/D)
\left\langle\frac{x}{1+e^{x+\delta}}\right\rangle_\delta dx
+\eta O_{K,M}\!\left(\frac{(\log D)^3}{D^3}\right)\notag\\
&=\frac{Kc_K\eta}{2D^2}\left(\frac{\pi^2}{6}+\Delta\right)
+\eta O_{K,M}\!\left(\frac{(\log D)^3}{D^3}\right).
\label{eq:app-Ddot-asymptotic}
\end{align}
The half-line integral is $\langle A_1\rangle_\delta/2$, by the symmetry of $\delta$. For a fixed class $1$, either member of the top pair contributes to contraction and source, giving $2(K-1)$ ordered roles. Their half-line kernel integrals are $1/2$ and $\mathcal B(\Delta)/2$, respectively. Hence
\begin{align}
\langle p_1(1-p_1)\rangle
&=\frac{(K-1)c_K}{D}+O_{K,M}\!\left(\frac{(\log D)^2}{D^2}\right),\label{eq:app-contraction-expansion}\\
\langle g_1^2\rangle
&=\frac{(K-1)c_K}{D}\mathcal B(\Delta)
+O_{K,M}\!\left(\frac{(\log D)^2}{D^2}\right).\label{eq:app-source-expansion}
\end{align}
Inserting these into the exact source--contraction equation \cref{eq:app-Delta-source-contraction} yields
\begin{align}
\dot\Delta&=\frac{Kc_K}{D}
\left[\frac{\eta^2}{b}\mathcal B(\Delta)-2\eta\Delta\right]\notag\\
&\quad+O_{K,M}\!\left(
\frac{\eta\Delta(\log D)^2}{D^2}
+\frac{\eta^2(\log D)^2}{bD^2}\right).
\label{eq:app-Deltadot-asymptotic}
\end{align}
The factors $\eta\Delta$ and $\eta^2/b$ in this remainder are essential under annealing. Estimating $\dot Q_{\mathrm{eff}}$ and $2D\dot D$ separately and then subtracting would obscure the exact factor $\Delta$ in the deterministic part. For completeness, adding $2D\dot D$ back to $\dot\Delta$ gives the centered-norm expansion,
\begin{align}
\dot Q_{\mathrm{eff}}
&=\frac{Kc_K}{D}\left[\eta\left(\frac{\pi^2}{6}-\Delta\right)
+\frac{\eta^2}{b}\mathcal B(\Delta)\right]\notag\\
&\quad+\eta O_{K,M}\!\left(\frac{(\log D)^3}{D^2}\right)
+\frac{\eta^2}{b}O_{K,M}\!\left(\frac{(\log D)^2}{D^2}\right).
\label{eq:app-Qeffdot-asymptotic}
\end{align}
Its drift coefficient also follows from the $A_2$ integral.

For the test loss, each ordered pair contributes the half-line loss kernel, whose integral is $\langle\mathcal L_0\rangle_\delta/2$. Thus
\begin{equation}
\mathcal L_g=\langle-\log p_y\rangle
=\frac{K(K-1)c_K}{2D}\left(\frac{\pi^2}{6}+\Delta\right)
+O_{K,M}\!\left(\frac{(\log D)^3}{D^2}\right).
\label{eq:app-testloss}
\end{equation}
These estimates prove the flow and loss statements of Theorem~\ref{thm:blexp}, with explicit remainders. In particular, they remain usable for vanishing learning rates and vanishing $\Delta$.

\subsubsection{Classification error}
\label{app:error-and-loss}

Misclassification requires a separate small-noise estimate, since its layer has width $\sqrt\Delta/D$, rather than $1/D$. For $D>0$, division of all logits by $D$ leaves their ordering unchanged. Writing $\sigma_\perp=\sqrt\Delta/D$,
\[
\epsilon_g=\Pr\!\left[\argmax_a(u_a+\sigma_\perp z_a)\ne\argmax_a u_a\right].
\]
The event that the teacher runner-up overtakes the teacher winner is contained in the error event. Its probability is
\begin{align*}
\int_0^\infty f_G(g)\Phi\!\left(-\frac{g}{\sqrt2\sigma_\perp}\right)dg
&=\sigma_\perp f_G(0)\int_0^\infty\Phi(-x/\sqrt2)\,dx+O_K(\sigma_\perp^2)\\
&=\frac{K(K-1)c_K}{\sqrt\pi}\,\sigma_\perp+O_K(\sigma_\perp^2).
\end{align*}
The remainder follows from the bounded derivative of $f_G$ and the finite first moment of the Gaussian tail kernel.
If an error occurs without this runner-up crossing, some third class must overtake the winner. This requires $G+G_2\le\sigma_\perp\max_{a,c}|z_a-z_c|$. The joint gap density is bounded, so the probability is at most $C_K\sigma_\perp^2\langle\max_{a,c}|z_a-z_c|^2\rangle=O_K(\sigma_\perp^2)$. For $K=2$ there is no such correction. We obtain
\begin{equation}
\epsilon_g=\Gamma_K\frac{\sqrt\Delta}{D}
+O_K\!\left(\frac{\Delta}{D^2}\right),
\qquad \Gamma_K=\frac{K(K-1)c_K}{\sqrt\pi}.
\label{eq:app-epsg}
\end{equation}
Equivalently, for an unordered boundary the local disagreement interval has length $|\delta|$, and $\langle|\delta|\rangle=2\sqrt{\Delta/\pi}$; summing $K(K-1)/2$ boundaries gives the same coefficient. Both the leading term and the remainder vanish at $\Delta=0$. This completes the proof of Theorem~\ref{thm:blexp}.

\subsection{Corollary \ref{cor:fixed}: fixed learning rate}
\label{app:fixed-rate-derivation}

For constant $\eta>0$, the limiting noise floor solves
\begin{equation}
2\Delta_*=\frac{\eta}{b}\mathcal B(\Delta_*).
\label{eq:app-Delta-star}
\end{equation}
To establish uniqueness and attraction, note that
\begin{equation}
\mathcal B(0)=2\log2-1>0,\qquad
\mathcal B'(\Delta)=\frac12\int Dz\,\operatorname{sech}^2\!\left(\sqrt{\frac\Delta2}z\right).
\label{eq:app-B-properties}
\end{equation}
The derivative is positive and strictly decreasing, and $\mathcal B(\Delta)=O(1+\sqrt\Delta)$. Thus $(\eta/b)\mathcal B(\Delta)-2\Delta$ is strictly concave, starts positive, and tends to $-\infty$. It has exactly one positive zero, with positive drift below and negative drift above it.

This sign information also controls the full trajectory, not just the reduced equation. By \cref{lem:entry}, $\Delta$ is bounded and $D\to\infty$. \cref{eq:app-Ddot-asymptotic} gives $\partial(D^3)/\partial\alpha=O(1)$, so $D^3=O(1+\alpha)$ and the slow time
\[
\tau(\alpha)=Kc_K\eta\int_{\alpha_1}^{\alpha}\frac{d\alpha'}{D(\alpha')}
\]
diverges, with $\alpha_1$ chosen after $D$ becomes positive. In this time,
\begin{equation}
\frac{d\Delta}{d\tau}=\frac{\eta}{b}\mathcal B(\Delta)-2\Delta
+O_{K,M}\!\left((\Delta+\eta/b)\frac{(\log D)^2}{D}\right).
\end{equation}
The perturbation vanishes uniformly on the bounded $\Delta$ interval. Outside any neighborhood of $\Delta_*$ it eventually cannot reverse the drift's sign. The trajectory therefore enters and remains in every such neighborhood, proving $\Delta\to\Delta_*$.

Multiplying \cref{eq:app-Ddot-asymptotic} by $3D^2$ gives
\[
\frac{\partial(D^3)}{\partial\alpha}=\frac{3Kc_K\eta}{2}\left(\frac{\pi^2}{6}+\Delta(\alpha)\right)
+\eta O_{K,M}\!\left(\frac{(\log D)^3}{D}\right).
\]
The error tends to zero. Integration and $\Delta\to\Delta_*$ imply
\begin{equation}
D^3(\alpha)\sim\frac{3Kc_K\eta}{2}\left(\frac{\pi^2}{6}+\Delta_*\right)\alpha.
\label{eq:app-D3-fixed}
\end{equation}
Consequently,
\begin{equation}
D\propto\alpha^{1/3},\qquad
Q_{\mathrm{eff}}=D^2+\Delta\propto\alpha^{2/3},\qquad
\epsilon_g\sim\frac{\Gamma_K\sqrt{\Delta_*}}{D}\propto\alpha^{-1/3},
\label{eq:app-fixed-scaling}
\end{equation}
and \cref{eq:app-testloss} gives $\mathcal L_g\propto\alpha^{-1/3}$. The relative loss remainder is $O((\log D)^3/D)$ and the relative error remainder is $O(\sqrt\Delta/D)$, so neither changes these laws. This proves Corollary~\ref{cor:fixed}.

The small-rate expansion is
\begin{equation}
\mathcal B(\Delta)=2\log2-1+\frac\Delta2+O(\Delta^2),
\qquad \Delta_*\sim\kappa\frac\eta b\quad\text{as }\eta/b\to0,
\label{eq:app-B-small}
\end{equation}
where $\kappa=(2\log2-1)/2$.
For a fixed non-small $\eta/b$, the root equation \cref{eq:app-Delta-star}, rather than this proportionality, determines the floor.
For completeness, if the online source is removed (population gradient flow), \cref{eq:app-Delta-source-contraction} instead gives $\Delta\to0$ and, for $\Delta>0$, $(\log\Delta)'\sim-2Kc_K\eta/D$. Since $D\propto\alpha^{1/3}$ still holds, $\log\Delta=-c\alpha^{2/3}+o(\alpha^{2/3})$ for some $c>0$. The error then decays faster than any power, whereas the loss retains one-third scaling.

\subsection{Corollary \ref{cor:sched}: power-law schedules}
\label{app:schedule-derivation}

\subsubsection{Tracking for \texorpdfstring{$0<\gamma<1$}{0 < gamma < 1}}

Consider the shifted power law used in the experiments,
\begin{equation}
\eta(\alpha)=\eta_0\left(1+\frac{\alpha}{\alpha_0}\right)^{-\gamma},
\qquad \eta_0,\alpha_0>0,\qquad 0<\gamma<1.
\end{equation}
Its accumulated learning time is
\begin{equation}
H(\alpha)=\frac{\eta_0\alpha_0}{1-\gamma}
\left[\left(1+\frac{\alpha}{\alpha_0}\right)^{1-\gamma}-1\right].
\label{eq:app-H-powerlaw}
\end{equation}
First determine the scales without assuming tracking. Bounded $\Delta$ and \cref{eq:app-Ddot-asymptotic} give $c_1\eta\le \partial(D^3)/\partial\alpha\le c_2\eta$ at late times, hence $D^3\propto H$. Define a slow time by $d\tau/d\alpha=Kc_K\eta/D$. Then $\tau\propto H^{2/3}\to\infty$. For any fixed $\varepsilon>0$, the drift in \cref{eq:app-Deltadot-asymptotic}, divided by $\eta/D$, is uniformly negative on $\Delta\in[\varepsilon,M]$ once $\eta$ is small. Its remainder tends to zero on that interval. The same trapping argument as above proves $\Delta\to0$. Returning to the alignment equation and integrating yields
\begin{equation}
D^3(\alpha)\sim\frac{Kc_K\pi^2}{4}H(\alpha).
\label{eq:app-D-schedule}
\end{equation}

The relaxation rate near the small-$\Delta$ root is $\lambda\sim2Kc_K\eta/D$. For this schedule,
\begin{equation}
\frac{|\eta'|/\eta}{\lambda}
\propto\frac{D}{\eta(\alpha+\alpha_0)}
\propto\alpha^{-2(1-\gamma)/3}\to0.
\label{eq:app-adiabatic-powerlaw}
\end{equation}
To show that this time-scale comparison actually implies tracking, set $v=b\Delta/\eta$. The factorized remainder in \cref{eq:app-Deltadot-asymptotic} gives
\begin{equation}
\frac{dv}{d\tau}=\mathcal B(\eta v/b)-2v
-\frac{D\eta'}{Kc_K\eta^2}v
+O_{K,M}\!\left((1+v)\frac{(\log D)^2}{D}\right).
\label{eq:app-tracking-equation}
\end{equation}
The last two coefficients vanish by \cref{eq:app-adiabatic-powerlaw} and $D\to\infty$. Keep the factor $1+v$ until boundedness is established: since $\mathcal B(\eta v/b)=\mathcal B(\Delta)\le\mathcal B(M)$, the right side is negative for all sufficiently large $v$, uniformly at late times. Thus $v$ eventually stays bounded. On this bounded interval the equation becomes $dv/d\tau=2\log2-1-2v+o(1)$, and the sign comparison gives
\begin{equation}
\Delta(\alpha)\sim\frac\kappa b\eta(\alpha),
\qquad \kappa=\frac{2\log2-1}{2}.
\label{eq:app-Delta-schedule}
\end{equation}
An unfactorized absolute error in $\dot\Delta$ would not justify this step as $\eta\to0$; \cref{eq:app-tracking-equation} shows why the retained remainders do.

Combining \cref{eq:app-Delta-schedule,eq:app-D-schedule,eq:app-epsg} gives
\begin{equation}
\epsilon_g(\alpha)\sim A_{K,b}\frac{\sqrt{\eta(\alpha)}}{H(\alpha)^{1/3}},
\qquad A_{K,b}=\Gamma_K\sqrt{\frac\kappa b}
\left(\frac{4}{Kc_K\pi^2}\right)^{1/3},
\label{eq:app-general-schedule}
\end{equation}
or, with the shifted power law inserted,
\begin{equation}
\epsilon_g(\alpha)\sim A_{K,b}(1-\gamma)^{1/3}\eta_0^{1/6}\alpha_0^{\gamma/6}
\alpha^{-(2+\gamma)/6}.
\label{eq:app-powerlaw-schedule}
\end{equation}
The relative classification-error remainder is $O(\sqrt\Delta/D)=o(1)$. Meanwhile,
\begin{equation}
\mathcal L_g(\alpha)\sim\frac{K(K-1)c_K\pi^2}{12D(\alpha)}
\propto H(\alpha)^{-1/3}\propto\alpha^{-(1-\gamma)/3},
\label{eq:app-testloss-scheduled}
\end{equation}
with relative remainder $O((\log D)^3/D)=o(1)$. This proves Corollary~\ref{cor:sched}.
The endpoint $\gamma=0$ has the one-third exponent by Corollary~\ref{cor:fixed}, but its floor must be obtained from \cref{eq:app-Delta-star}; the small-rate prefactor in \cref{eq:app-general-schedule} does not apply to arbitrary constant $\eta$.

\subsubsection{Borderline schedule: Proposition \ref{prop:borderline}}
\label{app:borderline-schedule}

For $\gamma=1$, write $\eta=A/(\alpha+\alpha_0)$ with $A=\eta_0\alpha_0$. The preceding argument up to \cref{eq:app-D-schedule} uses only $H\to\infty$ and $\eta\to0$, not tracking. It therefore still gives $\Delta\to0$ and
\begin{equation}
D^3(\alpha)\sim\frac{Kc_K\pi^2 A}{4}\log\alpha,
\qquad
\lambda(\alpha)\alpha\propto(\log\alpha)^{-1/3}\to0.
\label{eq:app-D-borderline}
\end{equation}
The variance now relaxes more slowly than the learning rate changes. To identify its actual behavior, set $t=\log(\alpha+\alpha_0)$. The exact equation \cref{eq:app-Delta-source-contraction} becomes
\begin{equation}
\begin{aligned}
\frac{d\Delta}{dt}&=-a(t)\Delta+q(t),\qquad
 a(t)=\frac{2KA}{K-1}\langle p_1(1-p_1)\rangle,\\
0\le q(t)&=\frac{KA^2e^{-t}}{b(K-1)}\langle g_1^2\rangle\le Ce^{-t}.
\end{aligned}
\label{eq:app-borderline-exact}
\end{equation}
By \cref{eq:app-contraction-expansion,eq:app-D-borderline},
\begin{equation}
\int_{t_0}^t a(s)\,ds=b_\Delta t^{2/3}+o(t^{2/3}),
\qquad
b_\Delta=\frac{3Kc_K A}{(Kc_K\pi^2 A/4)^{1/3}}>0.
\end{equation}
Variation of constants gives $\Delta(t)=e^{-\int_{t_0}^t a}\bigl[\Delta(t_0)+\int_{t_0}^t e^{\int_{t_0}^s a}q(s)\,ds\bigr]$. The bracket converges to a finite positive limit: its integrand is at most $\exp(Cs^{2/3}-s)$, and finite-batch SGD generates positive variance even from $\Delta(0)=0$. Consequently,
\begin{equation}
\log\Delta(\alpha)=-b_\Delta(\log\alpha)^{2/3}
+o\bigl((\log\alpha)^{2/3}\bigr),
\qquad \frac{\Delta(\alpha)}{\eta(\alpha)}\to\infty.
\label{eq:app-borderline-Delta}
\end{equation}
This proves Proposition~\ref{prop:borderline}. The remainder controls the logarithmic asymptotic, not a constant-prefactor equivalent for $\Delta$. \cref{eq:app-epsg} then shows that the error decays more slowly than $\alpha^{-p}$ for every $p>0$, whereas the loss decays as $(\log\alpha)^{-1/3}$. Thus $1/2$ is approached only as the limiting adiabatic exponent for $\gamma\uparrow1$, not at $\gamma=1$ itself.

For $\gamma>1$, $H(\infty)<\infty$. The exact bound $\dot D\le\eta\langle\max_a u_a-\min_a u_a\rangle/(K-1)$ makes $D$ converge to a finite limit. Since $\dot\Delta\le C_K\eta^2/b$, $\Delta$ remains bounded. The source and contraction in \cref{eq:app-Delta-source-contraction} are then integrable, so $\Delta$ also converges. The total contraction is finite, and hence cannot fully remove the positive variance generated by finite-batch SGD. Both loss and error consequently retain nonzero limiting values in this case.

\section{Binary warmup: smooth student for a hard teacher}
\label{app:binary-warmup}

This appendix records the simpler binary mechanism used as a warmup.  It is not needed for the $K$-class proof, but it shows that the same qualitative ingredients--a diverging norm, a shrinking angle, and an online-noise floor--also appear outside the softmax model.

Let $T,J\in\mathbb R^N$ with $T\cdot T/N=1$ and $\xi\sim\mathcal N(0,I_N)$, and define
\begin{equation}
    u=\frac{T\cdot\xi}{\sqrt N},
    \qquad
    t=\frac{J\cdot\xi}{\sqrt N}.
\end{equation}
The teacher label is $\tau(u)=\operatorname{sgn}(u)$ and the student output is
\begin{equation}
    g(t)=\operatorname{erf}\left(\frac{t}{\sqrt2}\right).
\end{equation}
For squared loss $\mathcal{L}=(\tau(u)-g(t))^2/2$, online gradient descent gives
\begin{equation}
    J^{\mu+1}=J^\mu+\frac{\eta}{\sqrt N}[\tau-g(t)]g'(t)\xi^\mu.
\end{equation}
With
\begin{equation}
    Q=\frac{J\cdot J}{N},
    \qquad
    \rho=\frac{J\cdot T}{N},
    \qquad
    R=\frac{\rho}{\sqrt Q},
\end{equation}
the fields $(u,t/\sqrt Q)$ are standard correlated Gaussians with correlation $R$.  The thermodynamic-limit flow is
\begin{align}
\frac{d\rho}{d\alpha}
&=\frac{2\eta}{\pi(Q+1)}\left[\sqrt{Q-\rho^2+1}-\frac{\rho}{\sqrt{2Q+1}}\right],
\label{eq:app-bin-rho}\\
\frac{dQ}{d\alpha}
&=\frac{4\eta}{\pi(Q+1)}\left[\frac{\rho}{\sqrt{Q-\rho^2+1}}-\frac{Q}{\sqrt{2Q+1}}\right]
\notag\\
&\quad+\frac{2\eta^2}{\pi^2\sqrt{2Q+1}}\left[\pi+2\arcsin\left(\frac{Q}{3Q+1}\right)-4\arcsin\left(\frac{\rho}{\sqrt{(3Q+1)(2(Q-\rho^2)+1)}}\right)\right].
\label{eq:app-bin-Q}
\end{align}
The explicit $\eta^2$ term is the variance of the online update.  Introducing $r=1-R$ gives
\begin{equation}
    \frac{dr}{d\alpha}=\frac{1-r}{2Q}\frac{dQ}{d\alpha}-\frac{1}{\sqrt Q}\frac{d\rho}{d\alpha}.
\label{eq:app-bin-r-identity}
\end{equation}
For $Q\gg1$, $r\ll1$, and $s=Qr=O(1)$, the large-$Q$ expansion has the form
\begin{align}
    \frac{dQ}{d\alpha}&=c(s,\eta)Q^{-1/2}+O(Q^{-3/2}),
\label{eq:app-bin-Q-asympt}\\
    \frac{dr}{d\alpha}&=r_3(s,\eta)Q^{-3/2}+O(Q^{-5/2}),
\label{eq:app-bin-r-asympt}
\end{align}
where
\begin{align}
    c(s,\eta)&=\frac{4\eta}{\pi}\left(\frac{1}{\sqrt{1+2s}}-\frac{1}{\sqrt2}\right)+\frac{\sqrt2}{\pi^2}\eta^2J(s),\\
    r_3(s,\eta)&=-\frac{4\eta s}{\pi\sqrt{1+2s}}+\frac{\eta^2}{\pi^2\sqrt2}J(s),\\
    J(s)&=\pi+2\arcsin\left(\frac13\right)-4\arcsin\left(\frac{1}{\sqrt{3(1+4s)}}\right).
\end{align}
A consistent power law requires $r_3(s_*,\eta)=0$, which fixes $s_*=\lim_{\alpha\to\infty}Qr$.  Then
\begin{equation}
    Q(\alpha)\sim\left[\frac32c(s_*,\eta)\alpha\right]^{2/3},
    \qquad
    r(\alpha)\sim\frac{s_*}{Q(\alpha)}\propto\alpha^{-2/3}.
\end{equation}
The binary classification error is
\begin{equation}
    \epsilon_g=\frac1\pi\arccos R=\frac1\pi\arccos(1-r)\sim\frac{\sqrt{2r}}{\pi},
\end{equation}
and hence $\epsilon_g\propto\alpha^{-1/3}$.  If the learning rate decays adiabatically, the same reduced structure gives a noise-floor relation $r\sim k\eta/Q$ and leads to $\epsilon_g\propto\alpha^{-(2+\gamma)/6}$ when $\eta\propto\alpha^{-\gamma}$ with $\gamma<1$. This binary calculation is a useful sanity check and demonstrates similar behavior beyond softmax and cross-entropy loss, but the main paper's results are the multiclass softmax boundary-layer formulas derived above.

\section{Numerical protocols and additional robustness checks}
\label{app:numerical-details}

\begin{table}[t]
\centering
\small
\caption{Comparison with the aligned-student theory of \citet{liu2026universal}.}
\label{tab:liu-comparison}
\begin{tabular}{@{}lll@{}}
\toprule
 & \citet{liu2026universal} & this work\\
\midrule
targets & peaked soft distributions & one-hot labels (zero-temperature limit)\\
dynamics & gradient flow, aligned-student ansatz & online SGD, rotational dynamics in $(D,\Delta)$\\
one-third law for & test cross-entropy loss (calibration) & classification error \emph{and} test loss\\
classification error & zero identically & $\epsilon_g\sim\Gamma_K\sqrt{\Delta}/D\propto\alpha^{-1/3}$, exact prefactor\\
gradient noise & left open & noise floor $2\Delta_*=(\eta/b)\,\mathcal B(\Delta_*)$\\
learning-rate schedules & loss-level reparametrization $\tau=\int\eta$ & error exponent $1/3\to1/2$ while loss slows\\
\bottomrule
\end{tabular}
\end{table}

This appendix summarizes the numerical settings underlying the main figures and records additional robustness checks.  The reported observables are the test misclassification rate $\epsilon_g$, the centered margin $D=R-S$, the residual variance $\Delta=Q_{\mathrm{eff}}-D^2$, as well as the test loss $\mathcal{L}_g$. The total computational cost for all simulations
reported in the manuscript was on the order of \(10^3\) CPU-hours.

For the two main teacher-student validation figures,
\cref{fig:fixed-eta-validation,fig:schedule-validation}, we used the
online \(K=3\) softmax teacher-student model at dimension \(N=500\).
The teacher vectors were chosen orthonormal with
\(T_a\cdot T_b/N=\delta_{ab}\), and the student weights were initialized
with independent entries \(J_{ai}\sim\mathcal N(0,1)\).  Each SGD update
used a fresh Gaussian input example, and time is reported in macroscopic
units \(\alpha=\mu/N\).  The curves were generated from six independent
random seeds. The plotted envelopes show the corresponding seed-to-seed
fluctuations around a representative trajectory.  For the fixed-learning-rate
comparison in \cref{fig:fixed-eta-validation}, the theoretical prediction
uses the full asymptotic prefactor: the boundary density \(c_K\) was evaluated
numerically, and for each fixed learning rate the residual variance floor
\(\Delta_*\) was obtained by numerically solving
\(2\Delta_*=\eta\,\mathcal B(\Delta_*)\), with \(\mathcal B\) defined in
\cref{eq:B-main}.  For the schedule experiment in
\cref{fig:schedule-validation}, the learning-rate family was
\(\eta(\alpha)=\eta_0(1+\alpha/\alpha_0)^{-\gamma}\), with
\(\eta_0=2.0\), \(\alpha_0=10.0\), and the plotted exponents
\(\gamma\in\{0,0.5,1\}\). The generalization error \(\epsilon_g\) was estimated by Monte Carlo evaluation
on \(M=10^5\) fresh test examples. Figures showing numerical values for the order parameters $D$ or $\Delta$ use the centered student and teacher vectors of class one to determine them.

\subsection{Number of classes: \texorpdfstring{$K$}{K}-dependence}
\label{app:K-dependence}

The main-text simulations focus on \(K=3\), while the asymptotic theory is stated
for fixed but arbitrary \(K\).  We therefore include an explicit \(K\)-sweep as a
check that the predicted fixed-\(K\) boundary-layer mechanism is not unique to the three-class case. For fixed learning rate and batch size $b=1$, the theory predicts that the
exponents are independent of \(K\),
\begin{align}
    D\propto \alpha^{1/3} \ ,
    \qquad
    \epsilon_g\propto \alpha^{-1/3} \ ,
\end{align}
while the prefactors depend on \(K\) through the boundary density \(c_K\) and the
classification-error constant
\begin{align}
    \Gamma_K=\frac{K(K-1)c_K}{\sqrt{\pi}} .
\end{align}
More explicitly,
\begin{align}
    D^3(\alpha)
    \sim
    \frac{3Kc_K}{2}\,
    \eta\left(\frac{\pi^2}{6}+\Delta_*\right)\alpha,
    \qquad
    \epsilon_g
    \sim
    \Gamma_K\frac{\sqrt{\Delta_*}}{D},
\end{align}
where the fixed-learning-rate noise floor \(\Delta_*\) is determined by
\(2\Delta_*=\eta\mathcal B(\Delta_*)\) and is independent of \(K\) at this leading
order.

\cref{fig:app-K-eta-sweeps} shows simulations for
\(K = 5,20,50,100\) at \(N=200\), with fixed learning rates
\(\eta = 1.0, 0.1, 0.01\) and \(\gamma=0\).  The fluctuation envelopes are computed over
six seeds.  Across all tested values of \(K\), the late-time trajectories
remain in excellent agreement with the predicted asymptotic structure. The centered
overlap follows the \(D\propto\alpha^{1/3}\) law, the residual variance approaches the
learning-rate-dependent floor, and the generalization error follows the resulting
\(\epsilon_g\propto \sqrt{\Delta}/D\) decay. 
Increasing \(K\) induces a later crossover into the
asymptotic regime.  This is expected since for larger \(K\), the
top teacher fields are closer and the pairwise boundary-layer approximation becomes
valid only once the centered margin given by $D$ is large compared to the typical extreme-value
scale, \(D\gg \sqrt{2\log K}\).  Thus larger \(K\) produces longer transients before
the fixed-\(K\) late-time theory becomes visible.

\begin{figure}[t]
    \centering
    \includegraphics[]{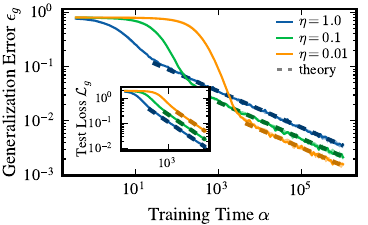}
    \includegraphics[]{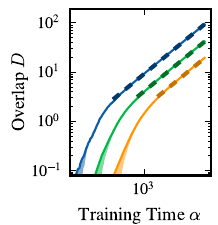}
    \includegraphics[]{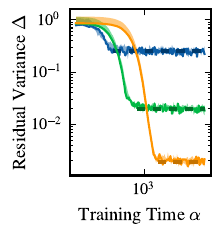}
    \includegraphics[]{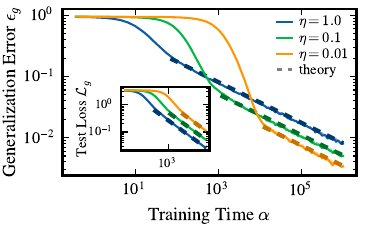}
    \includegraphics[]{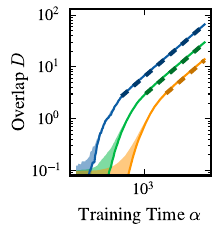}
    \includegraphics[]{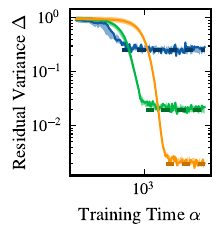}
    \includegraphics[]{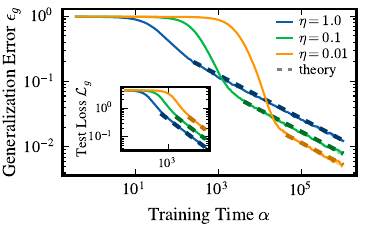}
    \includegraphics[]{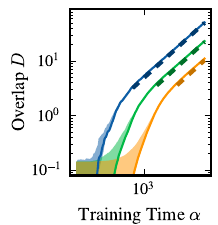}
    \includegraphics[]{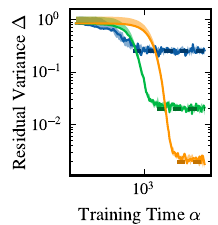}
    \includegraphics[]{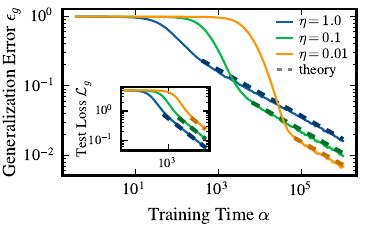}
    \includegraphics[]{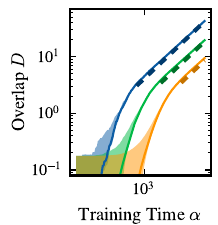}
    \includegraphics[]{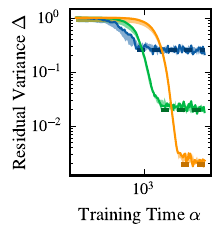}
    \addtocounter{figure}{-1}
    \caption{Dependence on the number of classes.
    Fixed-learning-rate simulations for \(K=5,20,50,100\) (increasing from top to bottom) at \(N=200\), with
    \(\eta=1.0,0.1,0.01\), \(\gamma=0\), and envelopes over six seeds.  For each value of \(K\), the panels show the generalization
    error \(\epsilon_g\), the centered student-teacher overlap \(D\), and the residual variance
    \(\Delta\).  The late-time behavior agrees with the fixed-\(K\) boundary-layer
    prediction: \(D\propto\alpha^{1/3}\), \(\Delta\) approaches the learning-rate
    noise floor, and \(\epsilon_g\propto\alpha^{-1/3}\).
    Increasing \(K\) delays the onset of the asymptotic regime.}
    \label{fig:app-K-eta-sweeps}
\end{figure}

\subsection{Batch size: \texorpdfstring{$b$}{b}-dependence}
\label{app:b-dependence}

At fixed learning rate, increasing the batch size suppresses the online-noise
term in \cref{eq:closure-lemma}, while leaving the drift of $D$ at a given state
unchanged.  The residual variance floor satisfies
\begin{equation}
    2\Delta_*=\frac{\eta}{b}\mathcal B(\Delta_*) \ ,
    \qquad
    \Delta_*\sim\frac{2\log 2-1}{2}\frac{\eta}{b}
    \quad \text{ for } \eta/b\to0 \ .
\end{equation}
Thus, for fixed finite $b$, the one-third exponents remain unchanged.  At small
$\eta/b$, the error prefactor decreases approximately as $b^{-1/2}$, whereas
$D$ and the test loss depend only weakly on $b$ through $\Delta_*$.

\begin{figure}[t]
    \centering
    \includegraphics[]{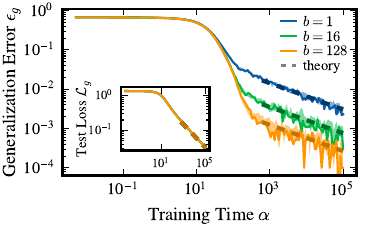}
    \includegraphics[]{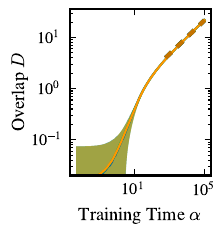}
    \includegraphics[]{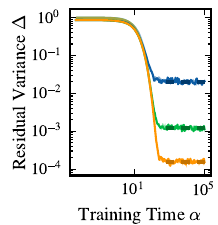}
    \caption{Batch-size dependence at fixed learning rate.  Left: generalization
    error, with test loss in the inset.  Middle: centered overlap $D$.
    Right: residual variance $\Delta$.  Solid curves show a single seed and shaded
    envelopes show the minimum--maximum over four seeds.  Dashed curves are the asymptotic
    predictions.  Larger batches reduce the noise floor and error prefactor
    without changing the late-time one-third exponents.}
    \label{fig:app-b-dependence}
\end{figure}

\cref{fig:app-b-dependence} compares $b\in\{1,16,128\}$ at $N=200$, $K=3$,
and $\eta=0.1$, using the orthonormal teachers and Gaussian student
initialization described above, with four seeds $0,1,2,3$.  Each update averages
the gradients of $b$ fresh Gaussian examples as in \cref{eq:online-update}.
Time $\alpha=\mu/N$ counts minibatch updates, so the number of training examples
is $bN\alpha$.  We train to $\alpha=10^5$ and evaluate at logarithmically spaced
checkpoints on $10^4$ independent Gaussian test inputs per seed, held fixed
across checkpoints and batch sizes.  The theoretical curves use the full
prefactors of \cref{cor:fixed,thm:blexp}, with $\Delta_*$ obtained by numerical
root finding and $\mathcal B$ evaluated using $5\times10^5$ standard Gaussian
samples.  Increasing $b$ lowers the observed residual variance and error,
while the overlap and loss curves nearly coincide, in agreement with the
small-$\eta/b$ prediction.

\subsection{Boundary-layer localization}
\label{app:boundary-decomposition}

To localize the examples responsible for error and alignment growth, let
$G=u_y-\max_{a\ne y}u_a$ be the top-two teacher-logit gap and
$\hat y=\argmax_a h_a$ the student prediction.  For the per-example cross-entropy
$\ell=-\log p_y$, the teacher-aligned negative loss gradient is
\begin{equation}
    -\frac{\partial\ell}{\partial D}
    =u_y-\sum_a p_a u_a\ge0 \ ,
    \qquad
    \dot D=-\frac{\eta}{K-1}\left\langle \frac{\partial\ell}{\partial D}\right\rangle \ .
    \label{eq:app-alignment-gradient}
\end{equation}
The second identity follows
from \cref{lem:closure}.  This diagnostic measures the per-example contribution to alignment
growth, not the loss itself or the full gradient norm.

We use the same isotropic teacher--student setup with $N=200$, $K=3$,
$\eta=0.5$, $b=1$, and one trajectory, saved at 20 approximately
logarithmically spaced checkpoints from $\alpha=10^2$ to $10^5$.
At each checkpoint, $2\times10^7$ fresh joint Gaussian teacher--student fields
are sampled.
The conditional profiles use 400 uniform gap
bins on $[0,0.6]$ for error and $[0,2.7]$ for the gradient.  At each checkpoint, we define the median teacher gaps for error and the
alignment-driving loss gradient, respectively, by
\begin{equation}
    \Pr\!\left(G\le G_{\epsilon_g}^{\rm median}(D)\mid\hat y\ne y\right)
    =\frac12,
    \qquad
    \frac{\left\langle -\frac{\partial\ell}{\partial D}\cdot\,\mathbf{1}\!\left\{
        G\le G_{\rm -\frac{\partial\ell}{\partial D}}^{\rm median}(D)\right\}\right\rangle}
        {\langle -\frac{\partial\ell}{\partial D}\rangle}
    =\frac12.
\end{equation}
Thus, the corresponding gap thresholds contain half of all classification
errors and half of the total alignment-driving gradient weight.

\begin{figure}[t]
    \centering
    \includegraphics[width=0.48\textwidth]{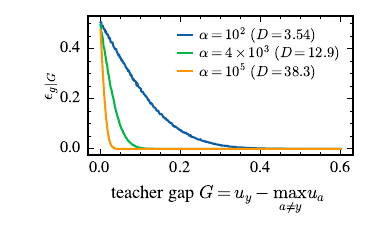}
    \hfill
    \includegraphics[width=0.48\textwidth]{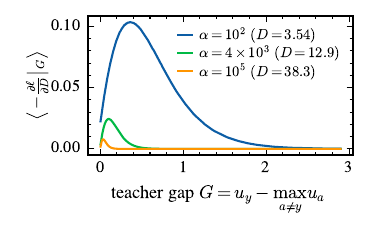}

    \includegraphics[width=0.48\textwidth]{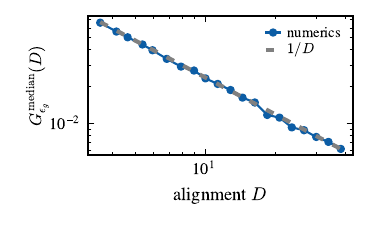}
    \hfill
    \includegraphics[width=0.48\textwidth]{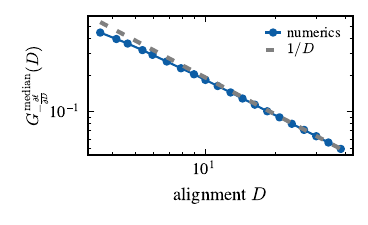}
    \caption{Direct boundary-layer diagnostics.  Top: error probability (left)
    and mean alignment-driving loss gradient (right), conditioned on the teacher
    gap, at $\alpha=10^2,\,4\times10^3,\,10^5$.  Bottom: corresponding median
    teacher gaps as functions of $D$.  The dashed $1/D$ guides are normalized
    at the final checkpoint. They test the scaling rather than a predicted
    prefactor.  Errors and alignment growth become increasingly localized near
    teacher decision boundaries.}
    \label{fig:app-boundary-decomposition}
\end{figure}

\cref{fig:app-boundary-decomposition} shows the conditional error
$\Pr(\hat y\ne y\mid G)$ and gradient $\left\langle -\frac{\partial\ell}{\partial D} \mid G\right\rangle$, together with
the gaps below which half of all errors or half of the total alignment-driving
weight is accumulated.  Both concentrate at smaller teacher gaps as $D$ grows.
The median gaps are consistent with $D^{-1}$ scaling: at fixed learning rate,
$\Delta$ stays near a nonzero floor, so both the $O(D^{-1})$ learning-signal
width and the narrower $O(\sqrt{\Delta}/D)$ error width shrink inversely with $D$.

\subsection{Correlated Gaussian inputs}
\label{app:correlated_gaussian}

For the correlated-Gaussian robustness check at $N=200$, the input covariance is diagonal with eigenvalues proportional to $i^{-\beta}$ and normalized so that the largest variance is one.  
Concretely, inputs were generated as
\(\xi_i=\sqrt{\lambda_i}\,x_i\), with \(x_i\sim\mathcal N(0,1)\) independently and
\begin{align}
    \lambda_i=\left(\frac{a}{a+i}\right)^\beta\ ,
    \qquad i=0,\ldots,N-1\ ,
\end{align}
with $a=10$ in the experiments, so that the largest variance has \(\lambda_0=1\) and increasing \(\beta\) produces a stronger power-law anisotropy.  The teacher vectors were sampled from independent standard-normal entries and then orthogonalized by Gram-Schmidt with respect to the normalized Euclidean inner product, followed by the normalization \(T_a\cdot T_b/N=\delta_{ab}\).  Thus the covariance spectrum is structured, but the teacher directions are otherwise random with respect to the covariance eigenbasis. The student weights were initialized with independent \(J_{ai}\sim\mathcal N(0,1)\), as in the isotropic simulations.
The labels are still generated by the teacher.  Increasing $\beta$ creates a structured spectrum and lengthens the transient. The observed late-time behavior remains consistent with the same boundary-layer exponent.

\subsection{Whitened feature experiments}
\label{app:vit_features}

The feature experiments use pretrained vision-transformer features as a non-Gaussian test bed \citep{dosovitskiy2021vit}.  Whitening removes the leading covariance structure, so the experiment probes whether non-Gaussian feature statistics destroy the boundary-layer pattern.  Teacher-generated labels are used in the main text because they avoid an early performance floor and allow the late-time regime to remain visible.  Real-label runs can reach a floor too early for a clean asymptotic fit, but they are useful as a practical diagnostic.

In these experiments we used CIFAR-5M images \citep{nakkiran2021deep} and extracted
fixed features with the pretrained ViT-B/16 model
\texttt{google/vit-base-patch16-224-in21k} \citep{dosovitskiy2021vit}.  Images were
passed through the ViT feature extractor and the pooled \texttt{[CLS]} representation
was used as the input feature vector.  For comparability with the \(K=3\)
teacher-student simulations, we restricted the dataset to three CIFAR classes:
airplane, automobile, and horse.  This class triple was chosen because, among the
tested three-class subsets, it gave one of the lowest apparent irreducible errors
when a linear classifier was trained on the ViT features, making it a favorable
case for observing a long late-time regime before a real-label floor is reached.
Before training the linear readout, we subtracted the empirical feature mean and
applied ZCA whitening: if \(X\) denotes the matrix of centered ViT features with
empirical covariance \(C=X^\top X/(n-1)\), we used the eigendecomposition
\(C=U\Lambda U^\top\) and transformed the features by
\(X\mapsto XU(\Lambda+\varepsilon I)^{-1/2}U^\top\), with a small numerical
regularizer \(\varepsilon=10^{-5}\).  The resulting whitened features have
approximately identity empirical covariance, but their distribution remains
strongly non-Gaussian. Teacher labels are the argmax predictions of a frozen bias-free linear classifier trained for one epoch on these features and real labels using cross-entropy SGD with learning rate \(5\times10^{-4}\) and batch size \(10\). The figures show envelopes over four seeds.

The readout architecture differs slightly from the analytically normalized
teacher-student model.  In the ViT-feature experiments we trained a linear layer, \(f(x)=Wx\), without the explicit \(1/\sqrt N\) factor in
the logit definition.  Consequently the runs used the default smaller PyTorch
linear-layer initialization and correspondingly smaller learning rates than in the
Gaussian teacher-student simulations.  Specifically, for input dimension \(N\),
\texttt{torch.nn.Linear} initializes weights from
the uniform distribution
\(\mathcal U(-N^{-1/2},N^{-1/2})\).  The readout was trained with cross-entropy
loss using online SGD without momentum, with batch size one. Time was again
reported in normalized units as the number of SGD steps divided by the feature
dimension.

\begin{figure}[t]
  \centering
  \includegraphics{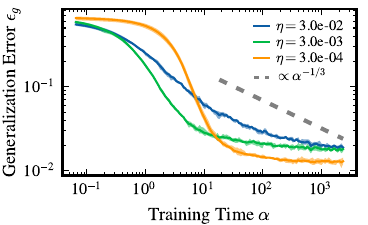}
  \caption{Whitened pretrained-feature experiment with real labels.  This run is included as a qualitative practical comparison to the teacher-label experiment in \cref{fig:vit-whitened}; the real-label performance floor is reached earlier, making it less suitable as a clean asymptotic test.}
  \label{fig:app-vit-comparison}
\end{figure}

The numerical evidence should be interpreted conservatively.  The asymptotic theory is for isotropic Gaussian inputs in the thermodynamic limit.  The correlated-input and whitened-feature experiments show that the mechanism can remain visible under controlled departures, but they do not constitute a theorem for arbitrary feature distributions or real-data classification.

\subsection{MNIST and Pythia experiments}
\label{app:mnist-pythia}

\begin{figure}[t]
  \centering
  \includegraphics[width=0.48\linewidth]{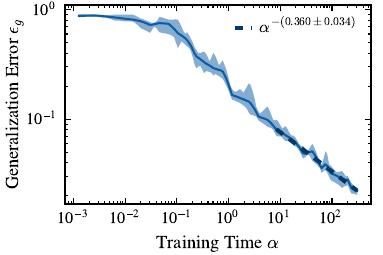}
  \hfill
  \includegraphics[width=0.48\linewidth]{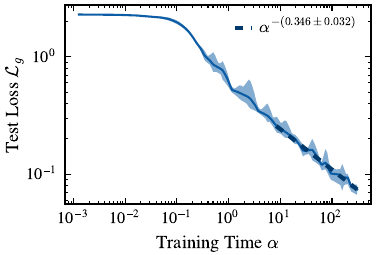}
  \caption{SGD on MNIST. The generalization error (left) and test loss (right) are compatible with one-third scaling. Curves are averaged over six seeds; the shaded envelopes span the seed trajectories. Dashed lines are fits over $\alpha\in[8.4,306]$. The exponent uncertainties shown in the panels are twice the covariance-based standard errors of the fits.}
  \label{fig:mnist-sgd-scaling}
\end{figure}

\paragraph{SGD on MNIST.}
We trained a $784$--$512$--$512$--$10$ ReLU MLP with cross-entropy and SGD at learning rate $0.01$, batch size one, and no momentum or weight decay for four epochs. The first epoch used the original images. The remaining three used random rotations up to $10^\circ$ and translations up to one pixel, reducing exact example reuse and bringing this finite-dataset setup closer to online SGD. \citet{liu2026universal} demonstrated a one-third loss law in an MNIST training run with full-batch Adam. We show that it persists for SGD. We report the mean and envelope over six seeds, with $\alpha$ equal to the number of updates divided by $784$. Fits of the across-seed means over $\alpha\in[8.4,306]$ show very good agreement of test loss with one-third scaling (\cref{fig:mnist-sgd-scaling}). Crucially, we also analyze test accuracy, reported through its complement, generalization error. Accuracy is the more direct performance metric for image classification, and the theory of Liu et al.\ makes no prediction for it. The observed error decay over this window is compatible with the one-third scaling our theory predicts.

\paragraph{Pythia next-token prediction.}
Pythia is a controlled suite of autoregressive transformer language models ranging from $70$M to $12$B parameters, with $154$ checkpoints per model released across training \citep{biderman2023pythia}. Following the checkpoint evaluation pipeline of \citet{liu2026universal}, we reproduced their Pythia loss curves and exponent estimates, confirming their results. However, changing either edge of the fit window by one checkpoint can change the exponent significantly, and considerably more than the covariance-based statistical fit error; we therefore report both contributions to the uncertainty below. Our contribution lies in extending this checkpoint analysis to token top-1 error. Evaluation uses the first $4{,}800$ documents of a pinned FineWeb stream \citep{penedo2024fineweb}, truncated to $1{,}024$ tokens, in batches of $12$.

Token top-1 error should be interpreted cautiously, because a text prefix can admit several plausible next tokens. It nevertheless probes a central distinction in LLM evaluation because raw probability magnitude and relative ordering of logits need not evolve together. Accuracy- and logit-rank-based scaling are established LLM questions: downstream top-1 error has been related to scaling laws \citep{gadre2025language}, while multiple-choice accuracy depends on margins against specific competitors and is not determined by correct-token loss alone \citep{schaeffer2025why}. Furthermore, decomposing cross-entropy loss reveals that a component based only on the rank of the correct logit compared to the largest one can scale more robustly than total cross-entropy \citep{yan2026scales}. Thus, better loss need not imply better task performance, and top-1 error directly probes the relative-logit dynamics central to our theory. In the Zipf model of \cref{app:zipf}, a student with fixed coverage recovers the boundary-layer $1/3$ exponent in its excess error once all represented sub-tasks have entered their late-time regime, despite their unequal frequencies. This motivates testing for the same exponent in Pythia, where extensive training at fixed model size could predominantly refine already represented patterns while limited coverage contributes an approximately constant error floor.

Pythia was trained with a learning-rate schedule, so raw training steps do not by themselves capture the varying update scale. \citet{liu2026universal} analyze the loss against what we define as $H(\alpha)=\int_0^\alpha \eta(\alpha')\,d\alpha'$, as considered before for learning rate schedules in \Cref{app:schedule-derivation}. For the error plots we consider a residual-aware training time $T_{\mathrm{res}}(\alpha)=H(\alpha)/\eta(\alpha)^{3/2}$. Under adiabatic tracking in our model, \cref{eq:app-general-schedule} reads $\epsilon_g\propto T_{\mathrm{res}}^{-1/3}$: the factor $\eta^{-3/2}$ incorporates the schedule-induced reduction of the residual variance $\Delta$. We fit $\epsilon_g=cT_{\mathrm{res}}^{-\beta}+\epsilon_{\mathrm{floor}}$, a power law plus a floor as in the loss analysis of \citet{liu2026universal} and as motivated by the sub-task model of \Cref{app:zipf}, over Pythia checkpoints $2{,}000$--$79{,}000$. For Pythia's adaptive optimizer, weight decay, and cosine schedule, this time coordinate is heuristic, not a consequence of Corollary~\ref{cor:sched} \citep{biderman2023pythia}.

The Pythia networks are deep transformers that differ substantially from a one-layer online teacher--student model, so exact exponent comparisons require cautious interpretation. The primary conclusion is that token top-1 error is well approximated by a power law over the selected fit window. The $12$B result is compatible with our one-third prediction, while the two smaller transformers have slightly steeper central fits. The reason for the steeper fits is not established.

\begin{figure}[t]
  \centering
  \includegraphics[width=0.48\linewidth]{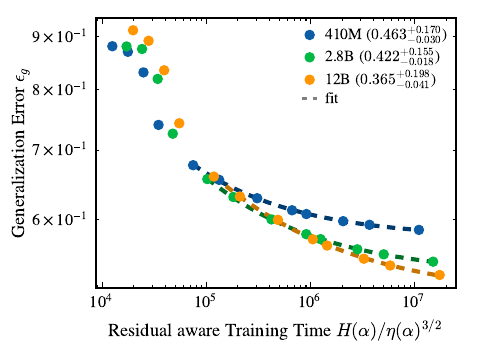}
  \hfill
  \includegraphics[width=0.48\linewidth]{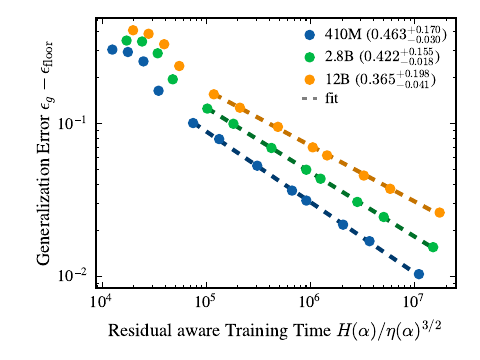}
  \caption{Pythia token top-1 error versus residual-aware training time. Left: fits to a power law plus a floor. Right: the same curves after subtracting the fitted floor. The central fit window spans checkpoints $2{,}000$--$79{,}000$. The uncertainties reported in the legends are asymmetric because, on each side, they sum the covariance-based statistical fit error with the largest exponent change obtained by shifting either fit-window edge by one checkpoint.}
  \label{fig:pythia-top1-scaling}
\end{figure}

\section{Aggregation over Zipf-distributed sub-tasks: a minimal coverage-limited model}
\label{app:zipf}

Real classification data can mix many sub-problems of very different frequency \citep{michaud2023quantization}, and
practical students represent only part of them. This appendix aggregates the
fixed-learning-rate law of \cref{cor:fixed} over Zipf-distributed
sub-tasks in a minimal routed-expert model. Its status differs from that of
\crefrange{sec:centered-dynamics}{sec:annealing}: conditioned on the routing,
every expert follows the dynamics of \cref{lem:closure} exactly, so the
per-sub-task exponent
agrees with the $1/3$ exponent derived in the main manuscript. The aggregation
below is a scaling-level calculation, not a
theorem. It is assembled from standard ingredients of quanta and template
models of scaling laws \citep{hutter2021learning, michaud2023quantization,
nam2024exactly}, which postulate the learning curve of each quantum; here that
curve is the boundary-layer law of the main text. All exponents stated below
refer to the regime in which every activated expert has reached its residual
noise floor $\Delta_*$ (\cref{cor:fixed}); before that, relaxation from above this floor can temporarily steepen an
expert's error decay without changing the leading loss scaling (see the
discussion following \cref{cor:fixed}).

\subsection{Model}
\label{app:zipf-model}

\paragraph{Data.}
An infinite dictionary of templates $k=1,2,\dots$ is sampled with Zipf weights
\begin{equation}
  p_k=\frac{k^{-\nu}}{\zeta(\nu)},\qquad \nu>1,
  \label{eq:zipf-weights}
\end{equation}
with normalization $\zeta(\nu)$.
A data point is a pair $x=(k,\xi)$: a template index and a local Gaussian
example $\xi\sim\mathcal N(0,I_N)$. 

\paragraph{Task.}
Template $k$ carries its own $K$-class sub-task with orthonormal teachers
\mbox{$T^{(1)}_k,\dots,T^{(K)}_k\in\mathbb R^N$}, $T^{(a)}_k\cdot T^{(b)}_k/N=\delta_{ab}$,
drawn independently across $k$; the label \mbox{of $(k,\xi)$} is
\mbox{$y=\arg\max_a T^{(a)}_k\cdot\xi/\sqrt N$}. Nothing about $T_k$ can be inferred
from other templates.

\paragraph{Student.}
A hard-routed mixture of linear experts: the router reads
$k$ and selects expert weights $J^{(1)}_k,\dots,J^{(K)}_k$, which produce logits
$t_a=J^{(a)}_k\cdot\xi/\sqrt N$, softmax probabilities, and the cross-entropy
loss of the main text. Training is online SGD on the stream with fixed learning rate
$\eta$ and for simplicity a batch size of one; a sample $(k,\xi,y)$ updates only the routed expert. Experts share no parameters, so, conditioned on the routing,
expert $k$ performs exactly the online dynamics of the main text, assumptions (A1)--(A7), in its own macroscopic time
\begin{equation}
  \alpha_k(t)=\frac{n_k(t)}{N}\simeq\frac{p_k\,t}{N} \ .
  \label{eq:zipf-local-time}
\end{equation}
Here,  $t$ counts stream samples and $n_k(t)$ the samples routed to $k$.
Consequently its test error is the base-model error at that time,
$\epsilon_k(t)=\epsilon_g(\alpha_k(t))$, and \cref{thm:blexp} and
\cref{cor:fixed} apply expert by expert. Only the
$\mathcal O(t^{1/\nu})$ experts of seen templates are ever updated.  

\paragraph{Three facts about Zipf sampling.}
(a) After $t$ draws, template $k$ has appeared $n_k\simeq p_k t$ times.
(b) Templates are seen only if $p_k t\gtrsim 1$, i.e.\ for ranks
$k\lesssim (t/\zeta(\nu))^{1/\nu}$; the number of distinct templates grows as
$t^{1/\nu}\ll t$ (Heaps' law).
(c) The probability mass of unseen templates (missing mass) is
$m(t)=\sum_{p_k t\lesssim 1}p_k\propto t^{-c_\nu}$ with
\begin{equation}
  c_\nu:=\frac{\nu-1}{\nu}=1-\frac1\nu\in(0,1) \ .
\end{equation}

\paragraph{Activation threshold.}
Expert $k$ enters the boundary-layer regime when $\alpha_k\approx\alpha_*$,
the entry scale of \cref{app:heuristic_entry_scale} (\cref{eq:heuritic_entry_scale}), i.e.\ at
stream time
\begin{equation}
  t_*(k)=\frac{N\alpha_*}{p_k}=N\alpha_*\,\zeta(\nu)\,k^{\nu},
  \qquad
  k_*(t)=\Big(\frac{t}{N\alpha_*\zeta(\nu)}\Big)^{1/\nu},
  \label{eq:zipf-threshold}
\end{equation}
where $k_*(t)$ is the rank of the activation frontier at time $t$. At scaling
level we take $\epsilon_k(t)=\mathcal O(1)$ for $t\lesssim t_*(k)$, more precisely of order the chance error
$\epsilon_0=1-1/K$, and
$\epsilon_k(t)=\epsilon_g(p_kt/N)\propto (p_kt/N)^{-1/3}$ for $t\gg t_*(k)$.

\subsection{Error decomposition and the one-pass law}
\label{app:zipf-onepass}

Independence of the sub-tasks gives
\begin{equation}
  \epsilon(t)=\sum_k p_k\,\epsilon_k(t)
  =\underbrace{\sum_{k>k_*(t)}p_k\,\epsilon_k(t)}_{\text{coverage}}
  +\underbrace{\sum_{k\le k_*(t)}p_k\,\epsilon_g\!\big(p_kt/N\big)}_{\text{refinement}} .
  \label{eq:zipf-decomposition}
\end{equation}
Coverage is a resource that only new data can buy; refinement is what training
buys on templates already in hand. At the stated scaling level, the coverage term is of order
$\epsilon_0\sum_{k>k_*}p_k\propto k_*^{-(\nu-1)}\propto t^{-c_\nu}$. In the
refinement term every activated expert contributes its own one-third law, and
the sum $\sum_{k\le k_*}p_k^{2/3}\propto\sum_{k\le k_*}k^{-2\nu/3}$ converges
if and only if $\nu>3/2$. Hence, for one pass over the stream,
\begin{equation}
  \epsilon(t)\;\propto\;
  \begin{cases}
    t^{-1/3}, & \nu>3/2\quad (c_\nu>1/3),\\[2pt]
    t^{-1/3}\log t, & \nu=3/2\quad (c_\nu=1/3),\\[2pt]
    t^{-c_\nu}, & 1<\nu<3/2\quad (c_\nu<1/3),
  \end{cases}
  \qquad\text{i.e.}\qquad \epsilon(t)\propto t^{-\min\{1/3,\,c_\nu\}}
  \label{eq:zipf-minrule}
\end{equation}
up to the logarithm at the marginal point. For light tails ($\nu>3/2$) the
frequent templates dominate and the boundary-layer exponent is the visible law.
For heavy tails ($\nu<3/2$) the activation frontier and the missing mass
contribute at the same order, and the $1/3$ is invisible in the total error even
though every activated expert individually follows it. The masking in
\cref{eq:zipf-minrule} is a property of the sum, not of its terms.

\subsection{Coverage-limited students and class-restricted observables}
\label{app:zipf-capacity}

Now let the student carry experts only for the $k_{\max}$ most frequent
templates and predict the remaining ones at chance. Its error has a static
floor set by coverage alone,
\begin{equation}
  \epsilon_{\rm floor}(k_{\max})=\epsilon_0\sum_{k>k_{\max}}p_k\propto k_{\max}^{-(\nu-1)},
  \label{eq:zipf-floor}
\end{equation}
and the excess error above the floor is the class-restricted error
$\epsilon_{\le k_{\max}}(t)=\sum_{k\le k_{\max}}p_k\,\epsilon_k(t)$, which
crosses over at $t_*(k_{\max})$:
\begin{align}
  \epsilon_{\le k_{\max}}(t)&\;\propto\; t^{-\min\{1/3,\,c_\nu\}},
  && t\ll t_*(k_{\max}),
  \label{eq:zipf-restricted-early}\\
  \epsilon_{\le k_{\max}}(t)&\;\sim\;
    \Gamma_K\sqrt{\Delta_*}\,
    \Big[\tfrac{3Kc_K}{2}\,\eta\big(\tfrac{\pi^2}{6}+\Delta_*\big)\Big]^{-1/3}
    \Big(\frac{t}{N}\Big)^{-1/3}\sum_{k\le k_{\max}}p_k^{2/3},
  && t\gg t_*(k_{\max}),
  \label{eq:zipf-restricted}
\end{align}
where \cref{eq:zipf-restricted} is \cref{cor:fixed} summed over the represented
experts, prefactor included, because all experts share $\eta$ and hence the same
noise floor $\Delta_*$. Below $t_*(k_{\max})$ the cutoff is invisible: the sum is
dominated by the frontier $k_*(t)<k_{\max}$ and by the not-yet-activated
templates below the cutoff. Above it the frequency tail is gone and the bare
boundary-layer exponent is the only time dependence left. For heavy tails
($\nu<3/2$) both branches are of order $k_{\max}^{-(\nu-1)}$ at the crossover,
comparable to the floor itself; for $\nu>3/2$ there is nothing to unmask.
Coverage limitation therefore does not remove the mechanism: it sets the last
stage of the approach to the floor, and the total error of a coverage-limited
student has the form ``power law plus floor'' used for the fits in
\cref{app:mnist-pythia}, with the power law becoming $t^{-1/3}$ once all
represented templates are activated.

This provides a possible interpretation of the Pythia analysis in \cref{app:mnist-pythia}. Each Pythia model is trained at fixed parameter count on a large text stream, with approximately $1.7\times10^{11}$ tokens processed by the upper end of our fit window \citep{biderman2023pythia}. Common linguistic patterns may therefore be well sampled even while finite model capacity limits the range of patterns that can be represented. In the present model, the corresponding regime requires an approximately fixed coverage cutoff $k_{\max}$ and sufficient training of all represented sub-tasks, $t\gg t_*(k_{\max})$: the student is capacity limited in coverage but data rich within its represented subset. Further training then refines predictions within that subset, while the unrepresented tail contributes a constant floor. This offers a mechanism for the power-law-plus-floor behavior considered in \cref{app:mnist-pythia}, with the schedule-dependent time coordinate used there providing a heuristic extension to Pythia. The aggregate fits alone do not establish that Pythia has reached this regime, and its fitted floor can also contain irreducible next-token uncertainty.

\end{document}